\documentclass[acmsmall]{acmart}
\usepackage[ruled,vlined]{algorithm2e}
\usepackage{comment}
\usepackage{subcaption}
\usepackage{multirow}
\usepackage{url}
\usepackage{relsize}
\usepackage{tabularx}
\usepackage{color}
\usepackage{ulem}
\usepackage{adjustbox}

\setcopyright{acmcopyright}
\acmJournal{TECS}
\acmYear{2022} \acmVolume{1} \acmNumber{1} \acmArticle{1} \acmMonth{1} \acmPrice{15.00}\acmDOI{10.1145/3542819}

\begin{document}

\title{Human Activity Recognition on Microcontrollers with Quantized and Adaptive Deep Neural Networks}

\author{Francesco Daghero}
\email{francesco.daghero@polito.it}
\orcid{0000-0001-9595-7216}
\affiliation{%
  \institution{Politecnico di Torino}
  \city{Turin}
  \country{Italy}
  \postcode{10129}
}

\author{Alessio Burrello}
\email{alessio.burrello@unibo.it}
\orcid{0000-0002-6215-8220}
\affiliation{%
  \institution{University of Bologna}
  \city{Bologna}
  \country{Italy}
  \postcode{40136}
}
\author{Chen Xie}
\email{chen.xie@polito.it}
\orcid{0000-0002-9225-3106}
\affiliation{%
  \institution{Politecnico di Torino}
  \city{Turin}
  \country{Italy}
  \postcode{10129}
}

\author{Marco Castellano}
\email{marco.castellano@st.com}
\affiliation{%
  \institution{STMicroelectronics}
  \city{Cornaredo}
  \country{Italy}
  \postcode{20010}
}

\author{Luca Gandolfi}
\email{luca.gandolfi1@st.com}
\affiliation{%
  \institution{STMicroelectronics}
  \city{Cornaredo}
  \country{Italy}
  \postcode{20010}
}

\author{Andrea Calimera}
\email{andrea.calimera@polito.it}
\orcid{0000-0001-5881-3811}
\affiliation{%
  \institution{Politecnico di Torino}
  \city{Turin}
  \country{Italy}
  \postcode{10129}
}

\author{Enrico Macii}
\email{enrico.macii@polito.it}
\orcid{0000-0001-9046-5618}
\affiliation{%
  \institution{Politecnico di Torino}
  \city{Turin}
  \country{Italy}
  \postcode{10129}
}

\author{Massimo Poncino}
\email{massimo.poncino@polito.it}
\orcid{0000-0002-1369-9688}
\affiliation{%
  \institution{Politecnico di Torino}
  \city{Turin}
  \country{Italy}
  \postcode{10129}
}

\author{Daniele Jahier Pagliari}
\email{daniele.jahier@polito.it}
\orcid{0000-0002-2872-7071}
\affiliation{%
  \institution{Politecnico di Torino}
  \city{Turin}
  \country{Italy}
  \postcode{10129}
}

\renewcommand{\shortauthors}{Daghero et al.}

\begin{abstract}

Human Activity Recognition (HAR) based on inertial data is an increasingly diffused task on embedded devices, from smartphones to ultra low-power sensors.
Due to the high computational complexity of deep learning models, most embedded HAR systems are based on simple and not-so-accurate classic machine learning algorithms. 
This work bridges the gap between on-device HAR and deep learning, proposing a set of efficient 1-dimensional Convolutional Neural Networks (1D CNNs) that can be deployed on general purpose microcontrollers (MCUs). 
Our CNNs are obtained combining hyper-parameters optimization with sub-byte and mixed-precision quantization, in order to find good trade-offs between classification results and memory occupation. Moreover, we also leverage adaptive inference as an orthogonal optimization to tune the inference complexity at runtime based on the processed input, hence producing a more flexible HAR system.

With experiments on 4 datasets, and targeting an ultra-low-power RISC-V MCU, we show that: i) We are able to obtain a rich set of Pareto-optimal CNNs for HAR, spanning more than 1 order of magnitude in terms of memory, latency, and energy consumption; ii) Thanks to adaptive inference, we can derive $>$ 20 runtime operating modes starting from a single CNN, differing by up to 10\% in classification scores and by more than 3x in inference complexity, with a limited memory overhead;
iii) On 3 of the 4 benchmarks, we outperform all previous deep learning methods, while reducing the memory occupation by more than 100x. The few methods that obtain better performance (both shallow or deep) are not compatible with MCU deployment.
iv) All our CNNs are compatible with real-time on-device HAR, achieving an inference latency that ranges between 9$\mu$s and 16ms. Their memory occupation varies in 0.05-23.17 kB, and their energy consumption in 0.05$\mu$J and 61.59$\mu$J, allowing years of continuous operation on a small battery supply.

\end{abstract}

\begin{CCSXML}
<ccs2012>
 <concept>
  <concept_id>10010147.10010257</concept_id>
  <concept_desc>Computing methodologies~Machine learning</concept_desc>
  <concept_significance>500</concept_significance>
  </concept>
 <concept>
  <concept_id>10010583.10010662.10010674</concept_id>
  <concept_desc>Hardware~Power estimation and optimization</concept_desc>
  <concept_significance>500</concept_significance>
  </concept>
 <concept>
  <concept_id>10010520.10010553.10010562</concept_id>
  <concept_desc>Computer systems organization~Embedded systems</concept_desc>
  <concept_significance>500</concept_significance>
 </concept>
 <concept>
  <concept_id>10010520.10010553.10010562.10010564</concept_id>
  <concept_desc>Computer systems organization~Embedded software</concept_desc>
  <concept_significance>500</concept_significance>
  </concept>
</ccs2012>
\end{CCSXML}

\ccsdesc[500]{Computing methodologies~Machine learning}
\ccsdesc[500]{Hardware~Power estimation and optimization}
\ccsdesc[500]{Computer systems organization~Embedded systems}
\ccsdesc[500]{Computer systems organization~Embedded software}

\keywords{quantized neural networks, mixed precision, adaptive neural networks, human activity recognition, edge computing, energy efficiency}

\maketitle

\section{Introduction}\label{sec:introduction}
\begin{figure}
    \centering
    \includegraphics[width=0.9\linewidth]{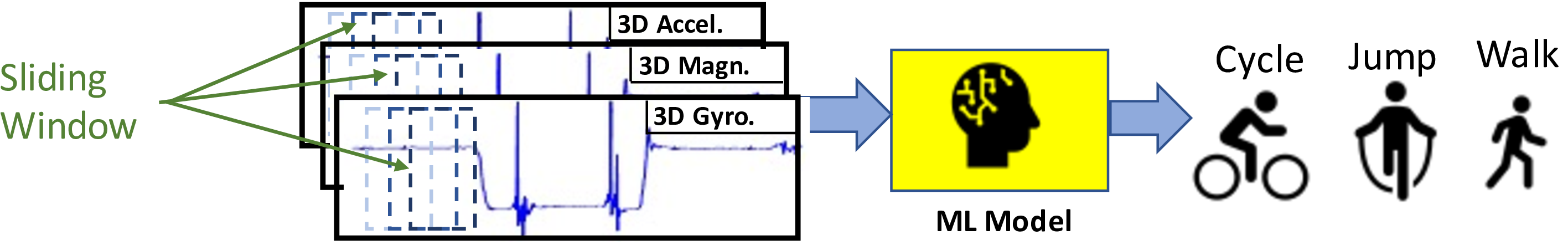}
    \caption{High-level overview of ML applied to a HAR task. The collected data are first windowed, then fed to the model. The latter will then try to predict the activity that was performed during the input window.}
    \label{fig:har_ml}
\end{figure}
Human Activity Recognition (HAR) based on Inertial Measurements Units (IMUs) is an increasingly common feature of smart connected devices such as smartwatches and fitness trackers.
The HAR problem consists in classifying the activity performed by a user, such as sitting, standing, walking, cycling, etc.
, based on a time-window of IMU readings, and is typically addressed with Machine Learning (ML) or Deep Learning (DL) algorithms~\cite{micucci2017unimib,comparison_hapt_dnn,comparison_wisdm_unimib_dnn,comparison_wisdm_rf}.
A typical ML flow for HAR is shown in Figure~\ref{fig:har_ml}, where the collected signals are first divided in fixed-size windows and then fed to a classifier.
Similarly to many other ML-based tasks in the Internet of Things (IoT) domain, HAR benefits from an on-device implementation, where the end-node autonomously produces the activity classification, without the need of offloading the computation to a cloud server.
On-device inference avoids the transmission of large amounts of (possibly privacy-sensitive) data to the cloud over energy hungry wireless channels, which may become slow or unavailable in bad connectivity areas~\cite{chen2019deep,daghero2020energy}.
Therefore, the benefits of this approach include better privacy, latency predictability and higher energy efficiency, where the latter is particularly critical since most devices for which HAR is relevant are battery-operated.

However, on-device HAR requires light-weight models, compatible with the tight memory requirements and relatively low clock frequencies of the ultra-low-power computational platforms embedded in fitness-tracker-like devices, which are typically based on Microcontrollers (MCUs).
For this reason, nowadays, most on-device HAR solutions are based on simple classic-ML algorithms, and in particular on tree-based models such as Decision Trees (DTs) and Random Forests (RFs).
These models are chosen due to their low inference complexity and simple software implementation, which enables real-time classification with limited memory usage~\cite{stsensor,fan2013human,balli2019human}.
In contrast, while there are several works demonstrating the theoretical effectiveness of DL for HAR in terms of classification results~\cite{comparison_hapt_dnn,comparison_wisdm_unimib_dnn,devita,bianchi2019iot}, these approaches are never actually deployed on-device, due to their prohibitive memory requirements and large number of operations for inference.

One of the most promising ways to make DL models compatible with edge devices is \textit{quantization}, an optimization based on reducing the precision used for storing data and performing calculations at prediction time.
Nowadays the standard approach to quantization is to replace 32bit floating point (fp32) operations either with lower precision floats (e.g. fp16) or with integer/fixed-point numbers (e.g. int8), significantly reducing both the memory and the computational requirements of the model.
In particular, int8 quantization causes negligible accuracy drops for most tasks, while reducing the memory footprint by up to 4x. 
Additionally, thanks to the broad support for 8-bit SIMD operations on general purpose hardware, the inference latency and energy consumption is also significantly lowered.

However, even 8-bit quantized DL models can still have memory footprints that are too large for ultra-low power MCUs.
Therefore, in recent years, multiple works have proposed \textit{sub-byte quantization} and \textit{mixed precision}~\cite{hubara,daghero2021ultracompact, yu2019tf,edmips, bruschi2020enabling}, as ways to further reduce the memory required to store the networks, thus enabling the deployment of larger models on extreme-edge devices. While sub-byte quantization simply refers to using less than 8-bit for all weight or activation values of the network, mixed precision allows different layers to use different bit-widths, thus avoiding the potential accuracy drops caused by a fixed low-bit-width solution, and creating a richer trade-off between accuracy and model size. 

Both fixed- and mixed-precision aggressive quantization schemes could be effective to deploy novel on-device HAR solutions based on DL, able to achieve higher accuracy than classical methods, with a comparable memory occupation. However, to the best of our knowledge, the usage of these techniques for HAR has not been studied extensively, especially targeting general purpose MCUs.

One limitation of quantization is that of being a \textit{static} optimization, applied to models \textit{before} they are deployed to the target hardware. Thus, once a model is deployed, its accuracy versus inference cost (e.g., latency or energy consumption) cannot be altered further. This limitation can be overcome thanks to so-called
\textit{dynamic} or \textit{adaptive} inference techniques,
which are based on tuning the computational effort of a DL model to the complexity of the processed input, and have recently emerged as orthogonal to standard static optimizations.

The intuition behind dynamic methods is that not all inputs are equally hard to process by a DL algorithm, and that ``easy'' inputs allow a more aggressive optimization. For instance, one could use either a smaller model, or better, activate only a part of the full model, when processing an easy sample. 
This, in turn, allows to vary the accuracy versus latency/energy consumption trade-off of the whole system \textit{at runtime}, by changing which inputs are considered ``easy'' and ``difficult'' depending on external triggers, such as the battery state of charge in an embedded node. Furthermore, if easy inputs are the majority, 
dynamic inference approaches can even outperform static ones, reducing the \textit{average} latency or energy consumption per input~\cite{park2015big,daghero2020energy}.
To our knowledge, similar to aggressive quantization, dynamic inference techniques have not yet been applied to HAR.

In this paper, we build upon these observations, and explore a new set of optimisations for on-device HAR based on DL. We propose to tackle the problem with compact 1-dimensional (1D) Convolutional Neural Networks (CNNs), and we analyze several orthogonal strategies to enable the deployment of these models on ultra-low-power microcontrollers.
In particular, in combination with an architectural search phase to derive good hyper-parameters settings, we explore sub-byte and mixed-precision quantization, as static optimizations targeting memory reduction. At the same time, we also explore the runtime tunability of the average latency and energy consumption of the networks by means of a dynamic inference. More in detail, the following are the main contributions of this work:

\begin{itemize}
\item We target four state-of-the-art HAR datasets, three public \cite{hapt, wisdm, micucci2017unimib} and one proprietary, with different characteristics in terms of data size, sampling frequency, number of activity classes, etc. 
\item On each dataset, we perform an extensive exploration of 1D CNNs architectures for HAR, deriving Pareto-optimal models spanning from less than 60 bytes to several kBs of memory, and correspondingly increasing accuracy.
\item Exploring sub-byte fixed- or mixed-precision quantization, we find networks that achieve up to 91\% memory reduction with negligible accuracy drops compared to 8-bit ones, showing the effectiveness of this type of optimization for HAR tasks.
\item By applying an adaptive inference mechanism on top of some of the models found in the previous steps, we show that the accuracy and computational effort can be tuned at runtime with a limited memory overhead ($<$ 15\%), yielding a final system that is up to 2.7x smaller with respect to a naive solution that deploys multiple independent static networks. Furthermore, on some datasets, adaptive inference reduces the average effort per input by up to 60\% with respect to a static model, with no accuracy drop.
\item We deploy some of the Pareto-optimal models derived from the above optimization on an ultra-low-power MCU platform~\cite{quentin}, based on an open-source, single-core RISC-V architecture~\cite{pulp}, in order to estimate latency and energy consumption.
\item We compare our results with state-of-the-art solutions based both on deep learning and on classical algorithms. We show that we are able to achieve higher or comparable or higher accuracy with lower memory requirements with respect to \textit{all} state-of-the-art models that fit on the target embedded device. We also achieve a new state-of-the-art accuracy among deep learning solutions for UniMiB-SHAR with 86.24\%/90.66\% balanced accuracy and F1 score, WISDM with 98.9\% F1 score and WALK with 95.74\% balanced accuracy.
\end{itemize}
The rest of the paper is organized as follows. Section~\ref{sec:background_and_related} provides some background and reviews the current state-of-the-art of ML approaches for HAR. 
The datasets and the hardware platform targeted in our work are described in Section~\ref{sec:materials}, whereas the explored optimizations are detailed in Section~\ref{sec:methods}.
Experimental results are discussed in Section~\ref{sec:results}, whereas Section~\ref{sec:conclusion} draws the conclusions of our work.

\section{Related Works}\label{sec:related}
\label{sec:background_and_related}
\begin{table}[ht]
\footnotesize
\resizebox{\textwidth}{!}{
\begin{tabular}{|lllllll|}
\hline
\textbf{Work} & \textbf{Dataset} & \textbf{Classes} & \textbf{Window} & \textbf{Features} & \textbf{Models} & \textbf{Score [\%]} \\ \hline 
\multicolumn{7}{l}{\textbf{Machine Learning (ML) methods}} \\  \hline 
\citet{micucci2017unimib} & UniMiB-SHAR$^*$  & 17 & 3 s & \begin{tabular}[c]{@{}l@{}}raw data\\ time-domain\end{tabular} & k-NN & 82.86 [bAcc.] \\\hline
\citet{hapt} & \begin{tabular}[c]{@{}l@{}}UCI HAPT$^*$  \\ REALDISP  \\ PAMAP2   \end{tabular} & \begin{tabular}[c]{@{}l@{}}8 \\ 33 \\ 12\end{tabular} & \begin{tabular}[c]{@{}l@{}} 2.56s \\ 6 s \\ 5.12 s \end{tabular} & time-domain & SVM &  \begin{tabular}[c]{@{}l@{}} 96.70 [Acc.]\\ 99.52 [Acc.]\\ 94.33 [Acc.] \end{tabular} \\\hline
\citet{wisdm} & WISDM$^*$  & 6 & 10 s & time-domain & ANN & 91.7 [Acc.] \\ \hline
\citet{mobiact} & \begin{tabular}[c]{@{}l@{}}Mobiact$^*$ \\ WISDM\end{tabular} & \begin{tabular}[c]{@{}l@{}}6 \\ 6\end{tabular} & 5 s  & \begin{tabular}[c]{@{}l@{}}time-domain \\ freq.-domain\end{tabular} & k-NN & \begin{tabular}[c]{@{}l@{}} 99.88 [Acc.] \\ 99.79 [Acc.]\end{tabular} \\ \hline
\citet{comparison_hapt_rf} & UCI HAPT & 12 & 5 s & \begin{tabular}[c]{@{}l@{}}time-domain \\ freq.-domain\end{tabular} & RF & 88 [Acc.] \\\hline
\citet{comparison_wisdm_rf} & WISDM & 6 & 10 s & time-domain & RF & 98.1 [F1] \\\hline
\citet{anguita2012human} & UCI HAR & 6 & 2.56 s & freq.-domain & SVM & 89.3 [Acc.] \\\hline
\citet{bayat2014study} & Self-collected & 6 & 1.28 s & \begin{tabular}[c]{@{}l@{}}time-domain\\ magnitude\end{tabular} & Ensemble & 91.15 [Acc.] \\\hline
\citet{attal2015physical} & Self-collected & 12 & 1 s & \begin{tabular}[c]{@{}l@{}}time-domain \\ freq.-domain\end{tabular} & KNN & 98.85 [F1] \\
\hline 
\multicolumn{7}{l}{\textbf{Deep Learning (DL) methods}} \\  \hline 
\citet{comparison_hapt_dnn} & \begin{tabular}[c]{@{}l@{}} UCI HAPT\\ MobiAct \end{tabular} & \begin{tabular}[c]{@{}l@{}} 8 \\ 11\end{tabular} & \begin{tabular}[c]{@{}l@{}}2.56 s\\ 2.56 s\end{tabular} & raw data & CNN-BiLSTM & \begin{tabular}[c]{@{}l@{}}97.98 [Acc.] \\ 96.16 [Acc.]\end{tabular} \\\hline
\citet{comparison_wisdm_unimib_dnn} & \begin{tabular}[c]{@{}l@{}}OPPORTUNITY\\ PAMAP2\\ UCI HAR\\ UniMiB-SHAR\\ WISDM\end{tabular} & \begin{tabular}[c]{@{}l@{}}18\\ 18\\ 6\\ 17\\ 6\end{tabular} & \begin{tabular}[c]{@{}l@{}}2.13 s \\ 5.12 s \\ 2.56 s \\ 3 s \\ 10 s\end{tabular} & raw data & \begin{tabular}[c]{@{}l@{}}CNN\\ lego-CNN \\ lego-CNN \\ CNN \\ lego-CNN\end{tabular} & \begin{tabular}[c]{@{}l@{}}86.1 [F1] \\ 91.4 [F1] \\ 96.90 [F1]\\ 77.8 [F1] \\ 98.8 [F1]\end{tabular} \\\hline
\citet{devita} & PAMAP2 & 5 & 0.48 s & raw data & CNN & 93.11 [Acc.] \\\hline
\citet{bianchi2019iot} & \begin{tabular}[c]{@{}l@{}}UCI HAR\\ Self-collected \end{tabular} & \begin{tabular}[c]{@{}l@{}} 6\\ 9\end{tabular} & \begin{tabular}[c]{@{}l@{}}2.56 s\\ n.a.\end{tabular} & raw data & CNN & \begin{tabular}[c]{@{}l@{}}92.5 [Acc.] \\ 97 [Acc.]\end{tabular} \\\hline
\citet{daghero2021ultracompact} & \begin{tabular}[c]{@{}l@{}}WALK\\ UniMiB-SHAR\end{tabular} & \begin{tabular}[c]{@{}l@{}}2\\ 17\end{tabular} & \begin{tabular}[c]{@{}l@{}}1.28 s\\ 3 s\end{tabular} & raw data & BNN & \begin{tabular}[c]{@{}l@{}}94.6 [bAcc.] \\ 68 [bAcc.]\end{tabular} \\\hline\hline
\textbf{This work} & \begin{tabular}[c]{@{}l@{}}UCI HAPT\\ WISDM\\ UniMiB-SHAR \\ WALK\end{tabular} & \begin{tabular}[c]{@{}l@{}}12\\ 6\\ 17 \\ 2\end{tabular} & \begin{tabular}[c]{@{}l@{}}5 s \\ 10 s \\ 3 s \\ 1.28 s\end{tabular} & raw data & CNN & \begin{tabular}[c]{@{}l@{}} 85.63 [Acc.] \\ 98.9/98.81 [F1/Acc.] \\ 86.24/90.66 [bAcc./F1] \\ 95.74 [bAcc.] \end{tabular} \\\hline
\multicolumn{7}{l}{$^*$ Work introducing the new dataset.}
\end{tabular}
}
\caption{State-of-the-art on ML-based HAR. Abbreviations: freq.: frequency; bAcc.: balanced Accuracy; Acc.: Accuracy; F1: F1 score. }\label{table:related}
\end{table}

ML approaches are becoming increasingly popular for HAR, leading to superior performance compared to classical algorithms (based on filtering and thresholding the signal) and being adopted in various applications, such as fall detection and health monitoring.
Table~\ref{table:related} summarizes the latest efforts in this field.
For each work, we report the benchmark dataset, the window input dimension, the features extracted (either in the frequency or time domain), the best model found, and its performance.

Most literature papers employ shallow ML algorithms such as Support Vector Machines (SVM), Random Forests (RF) and k-Nearest Neighbors (k-NN).
In \cite{anguita2012human}, a two-fold contribution is presented. First, the authors demonstrate the application of SVMs to a multi-class HAR dataset collected with smartphones. Further, they apply fixed-point arithmetic in the forward pass of the SVM to reduce its computation complexity and memory requirements while introducing a minimal drop of accuracy.
Focusing as well on smartphone-class devices, \cite{bayat2014study} proposes the usage of a low-pass filter pre-processing, a feature extraction phase and either one out of 6 ML classifiers or an ensemble of them. Specifically, the authors benchmark logistic model trees (LMT), logistic regression (LR), logit boost (LB), RF, SVM and shallow Artificial Neural Networks (ANN) on the target dataset. The classifiers are also benchmarked when combined in an ensemble, reaching 91.5\% accuracy (with a combination of ANN, LB and SVM), on their private dataset with 6 classes.
In \cite{micucci2017unimib}, the authors introduce a novel dataset composed of acceleration signals collected from smartphones. The authors benchmark this dataset with four classifiers: SVM, fully-connected ANNs, RFs and k-NN.
Four different tasks are considered, each with a different number of classes to recognize. Experiments are conducted using both the raw signals and magnitude-based features.
Results show that, for tasks related to classifying Activities of Daily Living (ADLs), k-NN obtains the highest balanced accuracy of 82.86\%.
Similarly, the work of~\cite{attal2015physical}, featuring a self-collected HAR dataset with 12 different activities, compares four supervised classifiers (k-NN, SVM, Gaussian Mixture Models and RF) and several unsupervised models, obtaining the best result of 98.85\% of F1 score when employing k-NN.
In~\cite{wisdm}, the authors benchmark their HAR dataset on three different classifiers, an ANN, a logistic regressor, and a decision tree. 
They demonstrate that the ANN slightly outperforms the other classifiers, although at the cost of higher computational complexity.
The authors of~\cite{hapt} propose a framework for HAR, benchmarking it on the UCI HAPT dataset with a reduced number of activities, the PAMAP2 and REALDISP datasets. 
Precisely, after windowing the collected data, they extract a set of features and perform the classification using an SVM.
In order to enhance the classifier, the authors introduce an additional module working on the SVM output probabilities. Specifically, they either prune the input window or smooth the output probabilities in case of classes with high misclassification rates. The extracted predictions are then buffered in an additional ``history'' module, so that the most likely activity among the current and previous predictions is selected as final label.
On the UCI HAPT dataset, with a 8-class variant of the task, they reach a 96.7\% accuracy.
In~\cite{mobiact}, the authors introduce a novel dataset featuring 13 activities, 9 daily life activities and 4 different type of fall motions. A classification of 6 activities is then performed on the newly introduced dataset and on a previously existing one (WISDM). 
The authors benchmark Decision Trees (DTs), k-NNs, ANNs and logistic regression, obtaining the best results with a k-NN classifier.
They achieve an accuracy of 99.88\% and 99.79\% respectively on their dataset and on WISDM.
Finally,~\cite{comparison_hapt_rf}, and ~\cite{comparison_wisdm_rf} benchmark RFs as hardware friendly models for HAR, reaching up to 88\% and 98.1\% accuracy on the UCI HAPT and WISDM datasets, respectively.

More recently, DL approaches for HAR have obtained state-of-the-art results on multiple datasets~\cite{hammerla2016deep}.
In \cite{bianchi2019iot}, the authors propose a solution based on Convolutional Neural Networks (CNNs) trained to recognize 9 different activities. The authors benchmark their network on a self-collected dataset as well as on the public UCI HAR dataset, obtaining up to 92.5\% accuracy.
With a partially binarized CNN, \cite{devita} obtains up to 93.67\% accuracy on the PAMAP2 dataset. The obtained architectures are deployed on a FPGA.
In~\cite{comparison_hapt_dnn}, the authors propose a hierarchical deep learning model composed by a CNN and a Bidirectional Long Short-Term Memory network (BiLSTM), benchmarking their solution on two public datasets: UCI HAPT and MobiAct.
On both datasets, and differently from our work, the postural transitions are clustered in two distinct groups, reducing the total number of classes to 8 for UCI HAPT and 11 for MobiAct.
Additionally, they perform an extensive ablation study on other classifier for HAR, including CNNs, k-NNs, SVMs and CNN-LSTMs.
They achieve 97.98\% average accuracy on the 8-class UCI HAPT and 96.16\% on MobiAct. 
Finally, the authors of~\cite{comparison_wisdm_unimib_dnn} propose the usage of smaller separable filters (called LEGO), thanks to which they are able to reduce the memory requirements of 1D CNN models significantly, with a minimal accuracy drop. These networks are benchmarked on 5 different state-of-the-art HAR datasets, comparing different filter parameters for classical 1D CNNs. They obtain an F1 score of up to  97.51 \% on the WISDM dataset and 74.46\% on UniMiB-SHAR.

While more accurate than shallow learners, these deep models are far more computationally complex and memory-hungry, making them difficult to deploy on constrained end-nodes. In \cite{agarwal2020lightweight}, the authors deploy a lightweight RNN, but consider the Raspberry Pi3 as their target, which is equipped with 1GB of RAM and consumes several Watts of active power. On the other hand, we target MCU-class devices, with < 1MB of memory and three orders of magnitude lower power consumption.
Similarly, the models of ~\cite{comparison_wisdm_unimib_dnn} and ~\cite{comparison_hapt_dnn} propose architectures with at least 0.3M floating point parameters (i.e., at least 1.2MB of model size), too large to deploy on MCUs.
\cite{devita} proposes aggressively optimized and small models, but they deploy them on FPGA rather than on general purpose MCUs. The additional freedom deriving from the possibility of customising the hardware configuration makes certain types of optimization much more effective than they would be on MCUs, which are, however, by far the most common type of processing device available on IoT devices for which HAR is relevant (wearables, fitness trackers, etc).

The only solution based on deep learning that explicitly targets MCU-class devices as a deployment target for HAR is presented in our previous work of~\cite{daghero2021ultracompact}, which leverages Binary Neural Networks (BNNs), i.e. NNs quantized to 1-bit precision.
These networks present several advantages in terms of memory and computational complexity, compared to other types of quantization~\cite{hubara}. Besides requiring up to 32$\times$ less memory with respect to a float model, they also completely drop arithmetic operations in favour of bit-wise ones. This, in turn, allows even general purpose CPUs to perform up to 32 operations in parallel, with significant energy efficiency improvements and speedups.
However, as shown in the following, limiting the precision to just 1-bit is often sub-optimal, and results either in unnecessarily large architectures (in terms of hyper-parameters configurations) or significant accuracy drops.

There also exist some commercial products implementing HAR on ultra-low-power hardware on the market, but currently they are based on shallow ML.
On example is the STM LSM6DSOX~\cite{stsensor}, a system-in-package featuring a 3D digital accelerometer and gyroscope, integrated with a digital Machine Learning Core (MLC), able to perform an on-chip classification of the data collected by the accelerometer to distinguish among several human activities.
The MLC core is currently constrained to use a small RF, limiting the maximum accuracy that can be achieved with this device.

\section{Materials}\label{sec:materials}

\subsection{Datasets}\label{sec:datasets}

We target four different HAR datasets, three public and one proprietary.
The first rationale for the selection of the three public benchmarks has been that they are widely-employed in the literature, thus allowing for an easy comparison with state-of-the-art algorithms, while also being well-documented and usable. Moreover, they are also diverse, both in terms of input data (type of sensor, sampling frequency, etc.) and complexity of the classification task (ranging from 6 to 17 classes).
Based on these criteria, we have chosen the following three popular datasets: UniMiB-SHAR~\cite{micucci2017unimib}, WISDM~\cite{wisdm} and UCI HAPT~\cite{hapt}.
The private dataset comes from an internal project, and has been selected due to its significantly different task complexity with respect to the other three.

For each of the four datasets, we use the same train/test split method and proportions as the original papers. All results are reported on the unseen test sets. Holdout sets taken as a random 25\% of the training set are additionally used to detect overfitting and perform early stopping during training. When overfitting is not spotted, holdout data are then re-added to the final training sets.

\textit{UniMiB-SHAR}~\cite{micucci2017unimib} features 11711 3-axial accelerometer records collected from the on-device sensors of Android phones.
The measurements are taken from 30 subjects, placing the device in the front pocket of the trousers. 
Acceleration values are sampled at 50 Hz, and the collection sessions are segmented in windows by the means of a ``peak-based'' approach. Specifically, whenever the acceleration magnitude is larger than 1.5 g (with g being the gravitational acceleration) at time $t$ and lower than 0 at time $t-1$, a 3s window is extracted around $t$.
The dataset features 17 activity classes, mixing 9 daily activities and 8 different falls. The authors propose several task variants depending on the considered classes (falls-only, daily-activities only, etc).
In this work, we benchmark our results on the variant with the largest number of classes, named AF-17, which features all the falls and the daily activities.
According to the original paper, we use 20\% of the data, randomly sampled, as test set.

\textit{UCI HAPT}~\cite{hapt} includes 958500 samples gathered from the gyroscope and accelerometer of Android phones. Data are collected at 50 Hz for 30 different subjects, and divided in windows of 5 seconds, i.e., 250 samples, with no overlap. The task consists of the recognition of 12 daily activities. We maintain the original per-subject train-test split suggested by the authors, with 30\% of the subjects in the test set. This dataset represents an interesting use-case due to the presence of gyroscope data.

\textit{WISDM}~\cite{wisdm} features 1.098 million sensor readings sampled at 20 Hz from tri-axial accelerometers. The data has been collected from 29 different subjects carrying the smartphone in their front pant legs pockets, while performing the following daily life activities: walk, jog, ascend stairs, descend stairs, sit and stand. Also in this case, we keep the data preprocessing unchanged from the one proposed by the dataset authors, dividing the samples in non overlapping 10s windows, which result in tri-axial input signals of dimension 200$\times$3.
For this dataset, a random 10\% of the samples is used as test set.

Finally, our private dataset, code-named WALK, contains 2387232 sensor readings sampled at 25 Hz from a tri-axial accelerometer, then divided in non-overlapping windows of dimension 32x3. The data has been collected with the sensor positioned, depending on the record, in one out of 9 positions, ranging from the hand of the subject to its backpack. During the recording, subjects performed one out of 7 possible activities: walking, running, climbing stairs, sitting on a bus or car, cycling, riding a bike or standing still.  In this case, the task is binary and it consists of determining whether the subject is ``walking'' or not (hence the name). For this dataset, we take a class-stratified 20\% of the samples as test data.

\subsection{Hardware Platform}\label{sec:nodes}

The great majority of IoT end-nodes are based on low-power microcontrollers (MCUs), whose main compute unit is a general purpose CPU, typically based on a RISC instruction set.
This is mainly due to the low cost and high programmability of these devices~\cite{pulp}, which makes them preferable to custom Application Specific Integrated Circuits (ASICs), potentially orders of magnitude more efficient, but whose design and manufacturing costs are only affordable for high-end, high-volume devices.
To bridge the efficiency gap, however, modern IoT processors are increasingly equipped with specialized architectural features, that benefit the execution of particular classes of applications~\cite{pulp}. This allows them to obtain high efficiency on particular application domains, while preserving generality and programmability.
Many of these specialized architectures are based on the open-source RISC-V Instruction Set Architecture (ISA)~\cite{risc_v_is_manual}, which occupies an increasingly large portion of the programmable IoT end-nodes' landscape~\cite{holler2019open}.
One of the main advantages of RISC-V is its flexibility and extensibility, thanks to which many companies and universities have proposed extensions to the basic ISA, leading to a plethora of chips specialized for different application domains, including edge ML and DL~\cite{quentin,sifive}.

Among those, our work focuses on the family of \textit{Parallel Ultra Low Power} (PULP) processors~\cite{pulp}, which includes both single- and multi-core chips optimized for low energy consumption, thanks to a combination of technological and architectural techniques.
Given the very low-power requirements and tight cost constraints of typical HAR wearable devices, we select one of the smallest architectures in the family, the single-core \textit{PULPissimo}.

\begin{figure}[ht]
    \centering
    \includegraphics[width=0.8\linewidth]{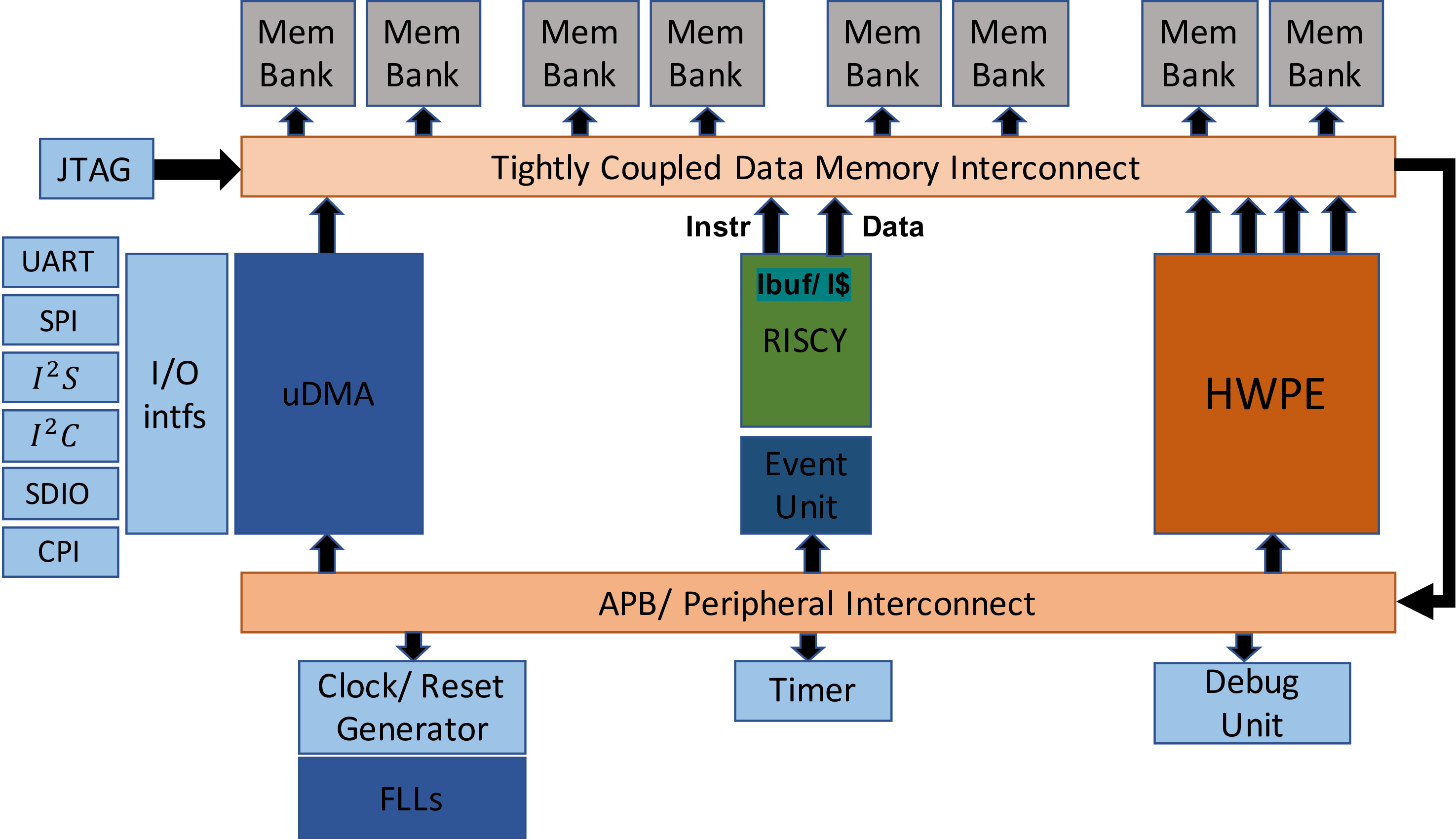}
    \caption{Schematic of the PULPissimo~\cite{pulp} architecture}
    \label{fig:pulpissimo}
\end{figure}

Fig. \ref{fig:pulpissimo} depicts a block diagram of PULPissimo, which is based on a RI5CY core~\cite{pulp}, with a 4-stage, in-order, single-issue pipeline. The core implements the \textit{RV32IMC} ISA, enhanced with domain-specific extensions for DSP, such as Single Instruction Multiple Data (SIMD) operations, hardware-loops, and loads/stores with index increment (\textit{XpulpV2} extension), that make it particularly suited to implement the linear algebra routines at the core of ML and DL inference efficiently~\cite{Burrello2021b}.
We target a 22nm implementation of this architecture, equipped with 520 kB of memory and reaching a maximum clock frequency of 938MHz~\cite{quentin}.
Given its characteristics, this device is representative of those found in low-cost and low-power IoT systems for which HAR is relevant, such as smart fitness trackers.

\section{Methods}\label{sec:methods}

The objective of this paper is to find optimized 1D CNN architectures for HAR through an extensive design space exploration.
We show that starting from a simple neural network template, and by combining multiple levels of optimization, it is possible to obtain a rich set of Pareto-optimal architectures, which achieve different trade-offs in terms of accuracy versus model size or accuracy versus latency/energy consumption, possibly tunable at runtime.

\begin{figure}[ht]
    \centering
    \includegraphics[width=\linewidth]{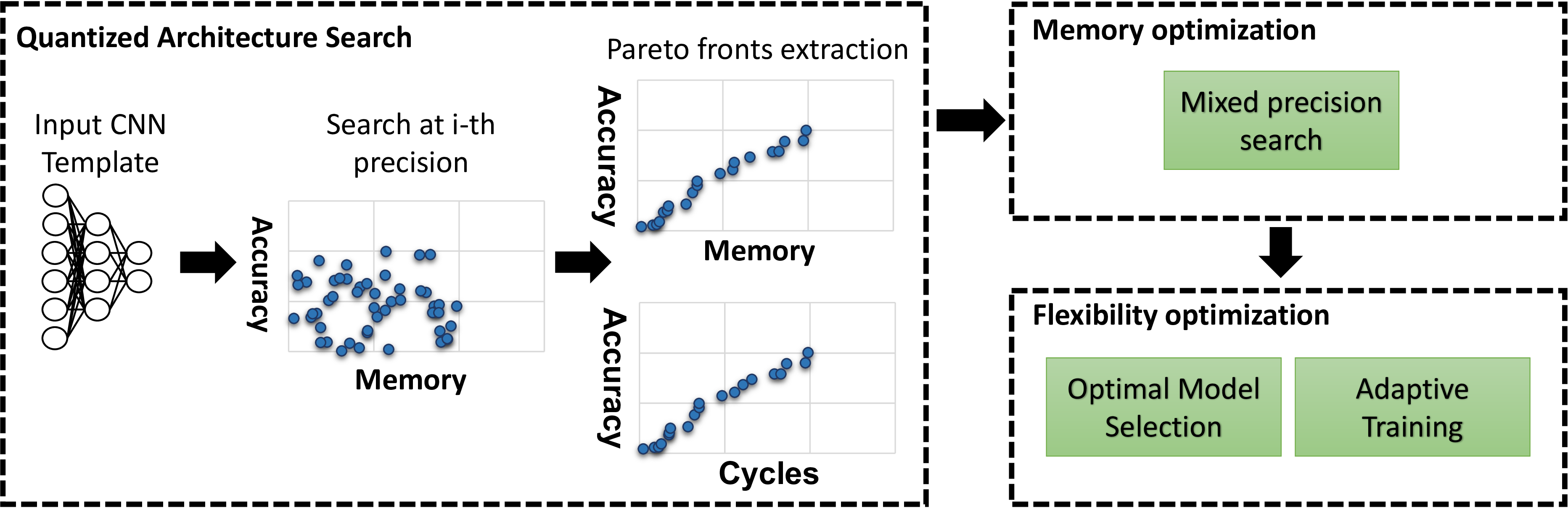}
    \caption{High-level scheme of the design space exploration flow followed in this work.}\label{fig:workflow}
\end{figure}

Figure~\ref{fig:workflow} depicts a high-level view of the flow that we adopt for this exploration. As shown, it is composed of three main blocks.
The first part, called \textit{Quantized Architecture Search}, consists in a hyper-parameters exploration based on grid search, and aimed at finding good 1D CNNs architectures. This search is repeated for different \textit{fixed-precision} quantizations, namely 8-bit, 4-bit, 2-bit and 1-bit.

Then, in the \textit{Memory Optimization} phase, we investigate \textit{mixed-precision} quantization in order to further optimize the accuracy versus memory occupation trade-off. 
Since an exhaustive search over all the possible combinations of bit-width assignments to the weights and activations of different CNN layers would be intractable, our framework employs a revisited version of a Neural Architecture Search (NAS) tool able to automatically select an appropriate precision for each tensor~\cite{edmips}. The NAS is applied to the results of the 8-bit architectural search, and its output are mixed-precision networks that include 1, 2, 4 and 8-bit layers.
With this part of our exploration, we aim at extending the results of~\cite{daghero2021ultracompact}, which only considered BNNs (i.e., 1-bit quantization), showing that ``intermediate'' bit-widths can yield superior results. Moreover, we also perform a comparison between fixed- and mixed-precision CNNs, showing that the latter are able to obtain superior performance on several memory ranges.

Lastly, for applications that require flexibility in terms of accuracy versus latency/energy consumption trade-off, we perform a further \textit{Flexibility Optimization} phase, in which we resort to adaptive inference to 
produce a model that, with a limited memory overhead ($<$ 15\%), supports a large number of different \textit{operating modes}, switchable at runtime by means of a single parameter change, with no re-training or re-deployment.
We then demonstrate that, depending on the dataset and on the ratio between easy and difficult inputs, the obtained adaptive networks
can also achieve significant energy/latency savings with respect to their static counterparts. 

Importantly, the two types of optimization proposed are orthogonal and fully independent.
For instance, it is in general possible to apply adaptive inference on standard 8-bit (or even floating point) networks.
In the rest of the section, each part of our exploration flow is described in detail.

\subsection{Quantized Architecture Search}\label{sec:grid_search}

\begin{figure}[ht]
    \centering
    \includegraphics[width=.9\linewidth]{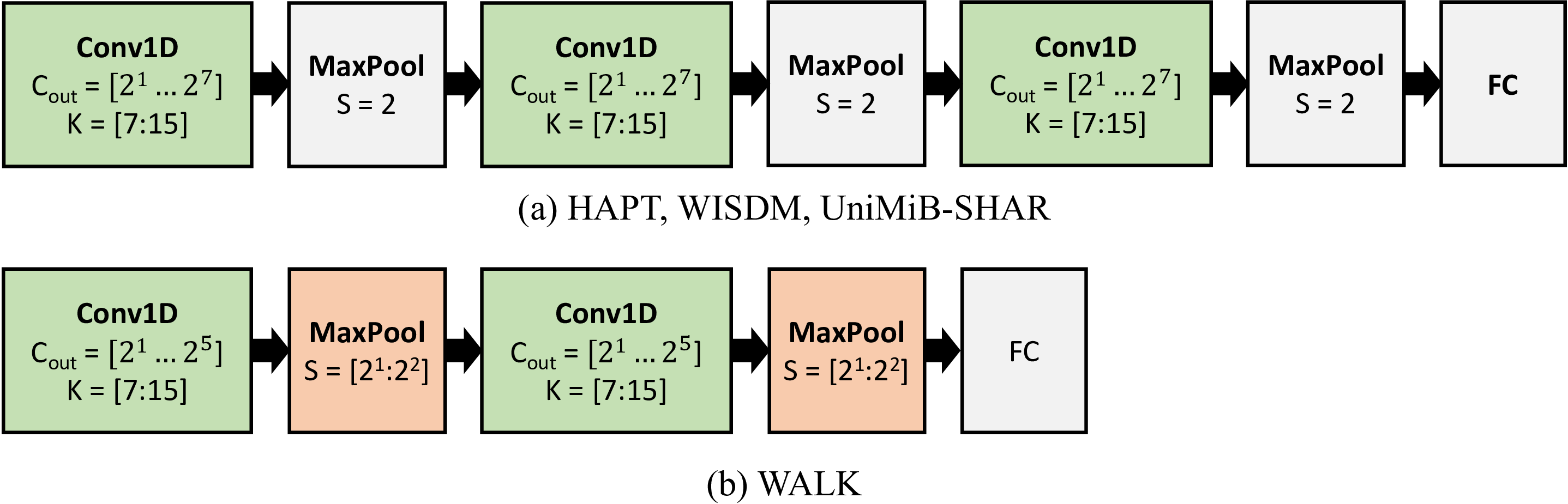}
    \caption{Overview of the input templates used for the quantized architecture search. The sub-caption refers to the datasets for which each template is used. The hyper-parameters of the layers in green and orange are optimized during the search, while grey layers are fixed. Orange layers can also be optionally removed. The granularity of the search is limited to power-of-two values for convolution output channels and pooling sizes. Abbreviations: Conv1D = 1-dimensional convolutional layer, MaxPool = 1-dimensional max pooling layer, FC = fully-connected layer, C$_{out}$ = number of output channels, K = kernel size, S = pooling size and stride.}
    \label{fig:grid_seach_net}
\end{figure}

Figure~\ref{fig:grid_seach_net} shows the 1D CNN templates used as starting point for our architectural exploration. We selected these templates empirically, with the goal of obtaining an acceptable accuracy while keeping the number of parameters low. We then performed an extensive search changing some of the key hyper-parameters of the layers depicted in green and orange. We kept the same search space for three open-source datasets targeted in our work (UCI HAPT, UniMiB-SHAR, and WISDM), as shown in Figure~\ref{fig:grid_seach_net}a, whereas we started from a smaller CNN template, shown in Figure~\ref{fig:grid_seach_net}b when targeting the WALK dataset. This choice is motivated by the easier binary classification problem addressed by WALK, which makes it useless to have more than two Convolutional layers or too many output channels in each layer.

Both templates are based on classic CNN architectures, such as LeNet~\cite{lenet}, appropriately modified to process the uni-dimensional multivariate time-series produced by accelerometers and gyroscopes used in HAR. They are composed of a sequence of alternating 1D convolutional layers (Conv1D) and max pooling layers (MaxPool), terminated by a single fully-connected layer (FC), with a number of output neurons equal to the number of classes of the corresponding dataset. Each Conv1D layer is followed by a batch normalization layer (BatchNorm), not shown in the figure for simplicity.

When targeting the three open datasets, our exploration varied the number of output channels ($C_{out}$) in each Conv1D layer, considering all power-of-two values between 2 and 128. Moreover, we also varied the convolution kernel size $K$, considering values 7 and 15. MaxPool layers were kept fixed, with pooling size ($S$) and stride both set at value 2.
For the WALK dataset, instead, the maximum $C_{out}$ of Conv1D layers was limited to 32, since in our initial experiments we found that larger values did not yield an improvement in accuracy. In addition, we expanded the search to also consider the hyper-parameters of MaxPool layers. This was done because, by shrinking the feature maps size, MaxPool is effective in further reducing the number of parameters of the final FC layer. Thus, we considered setting $S$ equal to 2 or 4, always keeping the stride equal to the pool size. At the same time, we also made MaxPool layers optional, in order to ensure that feature size shrinking did not limit the maximum accuracy achievable by our networks.

Given the small sizes of the templates of Figure~\ref{fig:grid_seach_net}, we explored hyper-parameters using a simple \textit{grid search} algorithm. This approach would be unfeasible for larger CNNs with 10s of layers, which would require the use of a smarter Neural Architecture Search (NAS) tool~\cite{Gordon2018,Risso2021a,Wan2020}, but when possible, it guarantees that all Pareto-optimal networks within the search space are found.

Moreover, in order to account for the interactions between network hyper-parameters and low-precision quantization, the grid search has been performed with Quantization-Aware Training (QAT)~\cite{Jacob2018}, and repeated at multiple bit-widths. 
Namely, we explored the hyper-parameters of fixed-precision CNNs (i.e., networks using the same bit-width for all layers, and for both weights and activations) representing data with 8-bit, 4-bit, 2-bit and 1-bit integers.
We selected these four bit-widths because they are those supported by the back-end inference library for our target MCU~\cite{bruschi2020enabling}.
For all bit-widths, we used the PArameterized Clipping acTivation (PACT) quantization algorithm, first proposed in~\cite{choi2018pact}. This algorithm was chosen due to its low-cost, full-integer, final resulting network, yet still accurate at low precision. Moreover, as for the set of precisions, PACT quantization was also chosen due to the compatibility with the open-source DNN deployment tool-chain available for the target hardware.

The rationale for repeating the grid search at each precision is that, if the data precision is decreased, a different setting of hyper-parameters might be needed to obtain Pareto-optimal networks. This is particularly true for BNNs (i.e., 1-bit networks), which typically need more channels per-layer to cope with the extreme precision reduction for weights and activations.

After the architectural exploration was completed, we 
extracted two sets of Pareto-optimal architectures for each bit-width. Precisely, we identified networks that are in the Pareto frontier either in terms of accuracy versus memory occupation, or in terms of accuracy versus number of cycles per inference. To this end, we used a C code template, based on the optimized inference toolchain of~\cite{Burrello2021b}, enhanced with the sub-byte precision kernels of~\cite{bruschi2020enabling}, to compile each CNN outputted by our grid search automatically. Each network was then simulated in the virtual platform of~\cite{gvsoc} to estimate the inference cycles. This allowed us to have a reliable ``proxy'' of the latency and energy consumption of the CNNs, rather than resorting to potentially inaccurate metrics such as the number of theoretical Multiply-and-Accumulate (MAC) operations.

\subsection{Memory-oriented Optimization with Mixed-precision CNNs}\label{sec:memory_opt}

Memory occupation is one of the main concerns for DL models deployment on MCUs, as detailed in Section~\ref{sec:background_and_related}.
This motivates the consideration of sub-byte quantization formats in our work, despite the fact that, with the notable exception of 1-bit BNNs, the latter typically do not yield latency and energy reductions on general purpose hardware, due to the lack of native hardware support for $<$8-bit operands, and the consequent need of unpacking/packing data before processing them~\cite{bruschi2020enabling}.

Previous work has shown that the optimal trade-off between memory occupation and accuracy is often not achieved quantizing the entire network at a single (fixed) precision, but rather resorting to a mixed-precision approach, in which each weights or activations \textit{tensor} in a DNN is allowed to use a different bit-width~\cite{bruschi2020enabling,edmips}.
Intuitively, a mixed-precision approach could assign larger bit-widths (up to 8-bit in our case) to layers for which a precise output representation is critical for the final accuracy of the whole model, while using a more aggressive quantization format for layers where output precision is less important.

Following this intuition, we consider this technique in our optimization of CNNs for HAR. The key problem in mixed-precision quantization is how to determine the assignment of bit-widths to the various tensors. In this case, an exhaustive grid search rapidly becomes unfeasible even for small network architectures. For instance, let us consider the template of Figure~\ref{fig:grid_seach_net}a, which includes three Conv1D layers and one FC layer. Evaluating all possible assignment combinations of 1-,2-,4- or 8-bit precision to weights and activations, for each of those four layers would involve $(4^2)^4 = 2^{16}$ independent training runs.
Furthermore, this number of trainings refers to a single setting of hyper-parameters, and the entire procedure would have to be repeated for different network architectures.

To tackle this problem, we employed an open-source tool named EdMIPS~\cite{edmips}. 
This tool frames the mixed-precision bit-width assignment problem as a \textit{Differentiable NAS} (DNAS)~\cite{Gordon2018,Risso2021a,Wan2020}, performing the optimization together with the training of the network weights, and in a comparable time. EdMIPS is significantly more light-weight than other mixed-precision search approaches based on reinforcement learning or evolutionary algorithms~\cite{mixed_nas_rl,mixed_nas_evo}.

\begin{figure}[ht]
    \centering
    \includegraphics[width=0.55\linewidth]{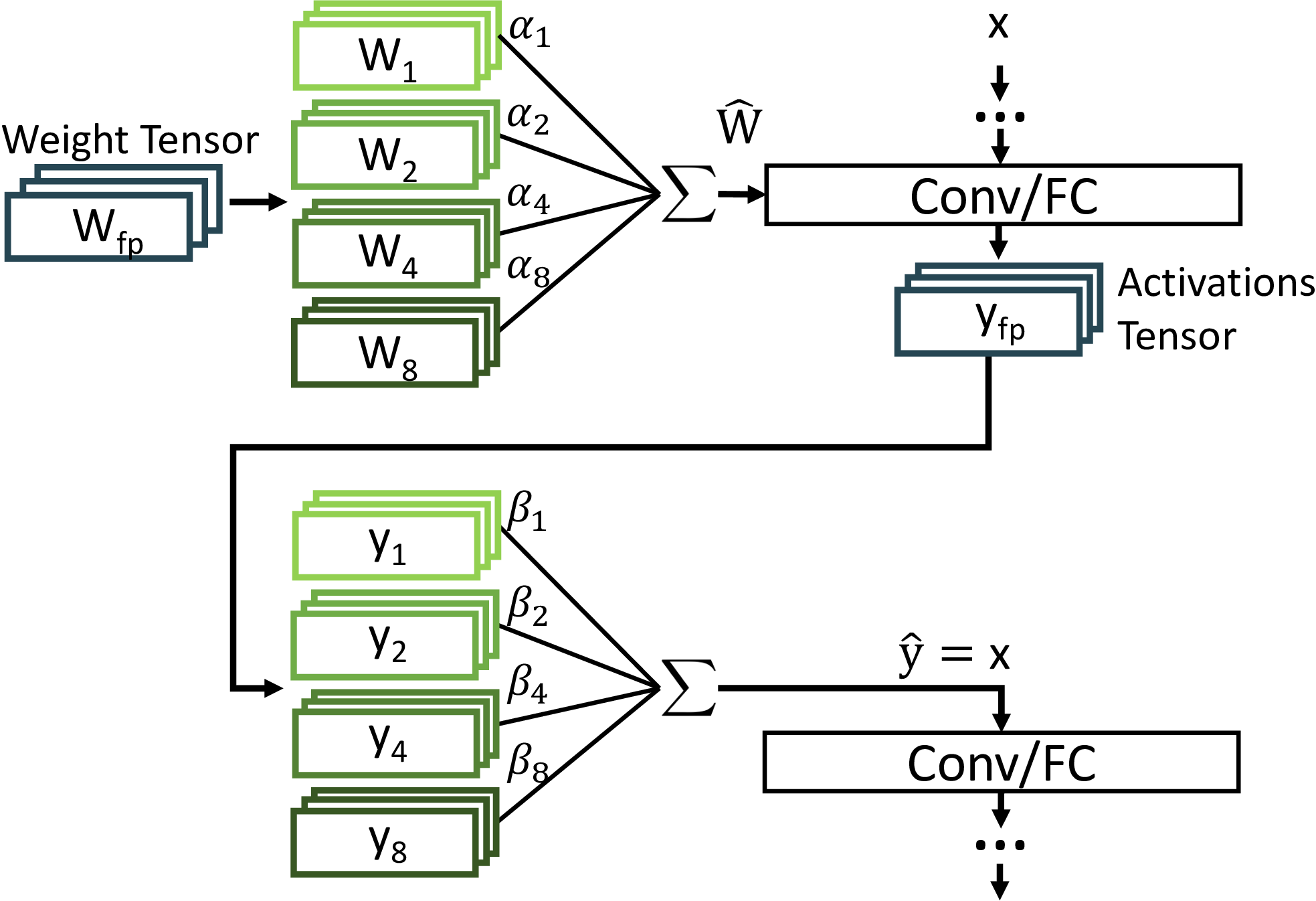}
    \caption{High-level view of the mechanism used in EdMIPS to search for the most suitable bitwidth for activations and weights.}
    \label{fig:edmips}
\end{figure}

A visual representation of how precision assignment can be made differentiable, allowing a gradient-based optimizer to determine an optimal bit-width for each tensor, is shown in Figure~\ref{fig:edmips}. The NAS uses two sets of trainable coefficients, called $\alpha$ and $\beta$, which are added to the standard parameters of the network. In each forward pass of the training loop, the quantization of both weights $W$ and output activations $y$ is simulated at all target bit-widths simultaneously. Taking weights as an example, different versions of the input floating point tensor $W_{fp}$ are created, called $W_1$, $W_2$, $W_4$, $W_8$, which store the values that would be obtained quantizing the weights to 1 2,4, and 8-bit integers respectively. Quantization is only simulated, in the sense that the actual values are still stored in floating point, but they are transformed with scaling/rounding/saturation, etc, (according to the chosen quantization algorithm) to mimic the effect of a lower-precision integer representation. The $\alpha$ coefficients, which are the outputs of a SoftMax operation (hence summing to 1), are then used to combine the different versions of the $W$ tensor as follows:
\begin{equation}
 \hat{W}=W_{1}\cdot \alpha_{1} + W_{2} \cdot \alpha_{2} + W_{4} \cdot \alpha_{4} + W_{8} \cdot \alpha_{8}
\end{equation}
Finally, the resulting $\hat{W}$ tensor is used in the corresponding Convolutional or FC layer. A similar procedure is also applied to the floating point layer outputs $y_{fp}$, using the $\beta$ coefficients, before feeding the resulting $\hat{y}$ as input to the following layer.

During training, a regularized loss composed of two terms is optimized:
\begin{equation}\label{eq:edmips_loss}
    \mathcal{L} = \mathcal{L}_{task}(W, \alpha, \beta) + \lambda \mathcal{L}_{cost}(\alpha, \beta)
\end{equation}
where $\mathcal{L}_{task}$ is the normal training loss (e.g., the categorical cross-entropy for a multi-class classification problem) and $\mathcal{L}_{cost}$ is a term that measures the expected computational cost of the network, e.g., in terms of total model size in bits, based on the current values of $\alpha$ and $\beta$. For example, the expected number of bits required for storing a layer's weights can be estimated as $(1 \cdot \alpha_1 + 2 \cdot \alpha_2 + 4 \cdot \alpha_4 + 8 \cdot \alpha_8) \cdot \|W_{fp}\|$, where $\|W_{fp}\|$ is the number of elements in $W_{fp}$.

Thanks to the loss function of (\ref{eq:edmips_loss}), the training converges to $\alpha$ and $\beta$ values that balance the accuracy loss induced by lower-precision quantization
with the associated lower computational cost. 
After convergence, the optimal precision for the weights and activations of a layer is simply selected by taking the argmax of $\alpha$ and $\beta$, respectively.
The regularization constant $\lambda$ allows to change the relative importance of accuracy and computational cost. Therefore, multiple trade-off points between the two metrics can be obtained by varying a single parameter, greatly reducing the search effort with respect to an exhaustive grid search.

Clearly, the one presented above is only a high-level overview of the functionality of EdMIPS, which involves many additional details and subtleties which have been skipped due to space limitations. Interested readers can refer to the original paper of~\cite{edmips} for more information.

We adapted EdMIPS, which was originally designed only for 2D computer vision CNNs, to work with 1D networks for HAR. Moreover, we also modified the tool to simulate the PACT quantization algorithm of~\cite{choi2018pact} rather than the original one proposed in~\cite{edmips}, which does not easily lend itself to a full-integer, hardware-friendly quantization.
We used the default cost metric proposed in~\cite{edmips}, and let the NAS optimize the precision of \textit{all} Conv1D and FC layers of our CNNs, selecting between 1, 2, 4 and 8-bit precision for both weights and activations. MaxPool layers have been set to use the same precision as the preceding convolutions.
The rationale for the choice of the set of precisions and quantization algorithm is the same explained in Section~\ref{sec:grid_search}.
We applied the mixed-precision NAS to each 8-bit network that was found Pareto-optimal in terms of accuracy versus memory, after the hyper-parameters exploration of Section~\ref{sec:grid_search}. From each of these starting points, we obtained \textit{multiple} mixed-precision networks by repeating the search varying $\lambda$. In particular, we selected 10 $\lambda$ values, from a minimum of 0.0001 up to 0.001, since values outside this range generally yielded networks either fully binarized or fully quantized at 8-bit.

\subsection{Flexibility-oriented Optimization with Input-adaptive Variable-width CNNs}\label{sec:adaptive}

As anticipated in Section~\ref{sec:introduction}, while optimizations such as quantization and binarization are fundamental to port deep learning on edge nodes, their limitation is that of being \textit{static}. That is, once a model is optimized,
its complexity versus accuracy trade-off cannot be altered further at runtime.
This is sub-optimal in many scenarios, as it does not allow to respond to changes in external conditions (e.g., the battery state-of-charge).
A naive way to obtain a variable accuracy versus complexity trade-off is to deploy \textit{multiple independent} neural networks on the target hardware and switch among them at runtime. However, this has an obvious drawback given by the huge memory overhead, since supporting N operating modes requires the storage of N different ML models. The rationale of input-adaptive techniques is to reduce this overhead by combining two mechanisms~\cite{daghero2020energy,survey_dynamic}.

\begin{figure}[ht]
    \centering
    \includegraphics[width=0.95\linewidth]{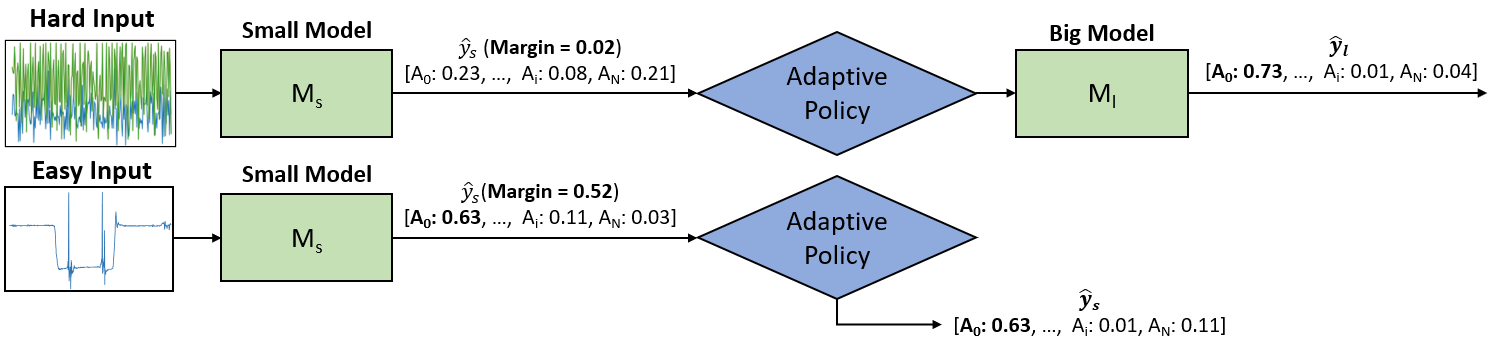}
    \caption{High-level overview of an adaptive inference approach in which computational complexity is tuned based on the input.} %
    \label{fig:adaptive_overview}
\end{figure}

First, the number of models is reduced from N to K, with 2 $\le$ K $<$ N. Each input is then assigned to one of K ``difficulty levels'' and processed with the corresponding model. By changing the policy that assigns difficulty levels to inputs, and consequently the frequency of usage of each of the K models, many intermediate operating modes ($\gg K$) can be obtained in terms of classification score versus average inference complexity.
Several variants of this technique have been proposed in literature, but most of them can be traced back to a common scheme, shown in Figure~\ref{fig:adaptive_overview} for $K=2$. Since assigning a ``difficulty-level'' to an input a priori (i.e., before processing it) is seldom possible, an incremental approach is used instead:
\begin{itemize}
    \item A simpler, less accurate model ($M_s$ in the figure) is executed first on each input.
    \item Then, the \textit{Adaptive Policy} block computes a metric of \textit{confidence}, based on the outputs of $M_s$.
    \item If the confidence is high enough, the input is deemed ``easy'', the output of $M_s$ is committed, and the inference ends. 
    \item Otherwise, a larger and more accurate model ($M_l$) is invoked on the same input, and its output is then used for the final classification.
\end{itemize}

The expected inference cost per input of this approach can be computed as follows:
\begin{equation}\label{eq:exp_cost}
    \overline{C} = C(M_s) + C(Policy) + (1 - \mathbb{P}_{e}) \cdot C(M_l)
\end{equation}
where $C(X)$ stands for ``cost of $X$'' (latency or energy depending on the objective) and $\mathbb{P}_{e}$ is the probability that an input is ``easy-enough'' to trigger early stopping after the execution of $M_s$.
Intermediate operating modes are obtained by changing the behavior of the Adaptive Policy block, which in turn influences $\mathbb{P}_e$.
The concept can also be naturally extended to more than two ``stages'' ($K>2$), although we do not consider that scenario in this work. 

The second mechanism used to reduce memory overheads is to let $M_s$ and $M_l$ \textit{share their parameters}. In fact, while early research proposed to implement the scheme of Figure~\ref{fig:adaptive_overview} with two independent neural networks of different size (obtaining so-called big-little systems)~\cite{park2015big}, this solution still incurs an almost 2x memory overhead compared to a static model.
Therefore, successive improvements have tried to \textit{overlap} the parameters of $M_s$ and $M_l$ in different ways.
In~\cite{teerapittayanon2016branchynet}, this is achieved generating $M_s$ from the initial layers of $M_l$, with an additional ``early-exit'' branch.
Similarly, in~\cite{tann2016runtime,yu2018slimmable, mullapudi2018hydranets,feature_boosting_suppression}, smaller models are obtained using a subset of the \textit{channels} (feature maps) of the largest one, whereas in~\cite{JahierPagliari2018a} they are obtained varying the quantization bit-width.
Other advanced adaptive inference mechanisms are described in~\cite{layerskipping,jahier2020sequence,daghero2020energy}. Importantly, many of these papers have shown that, if ``easy'' inputs are the majority at test time, input-adaptive systems can not only enable a higher flexibility in terms of operating modes, but also reach better trade-offs with respect to using multiple independent static networks.

\begin{figure}[ht]
    \centering
    \includegraphics[width=0.7\linewidth]{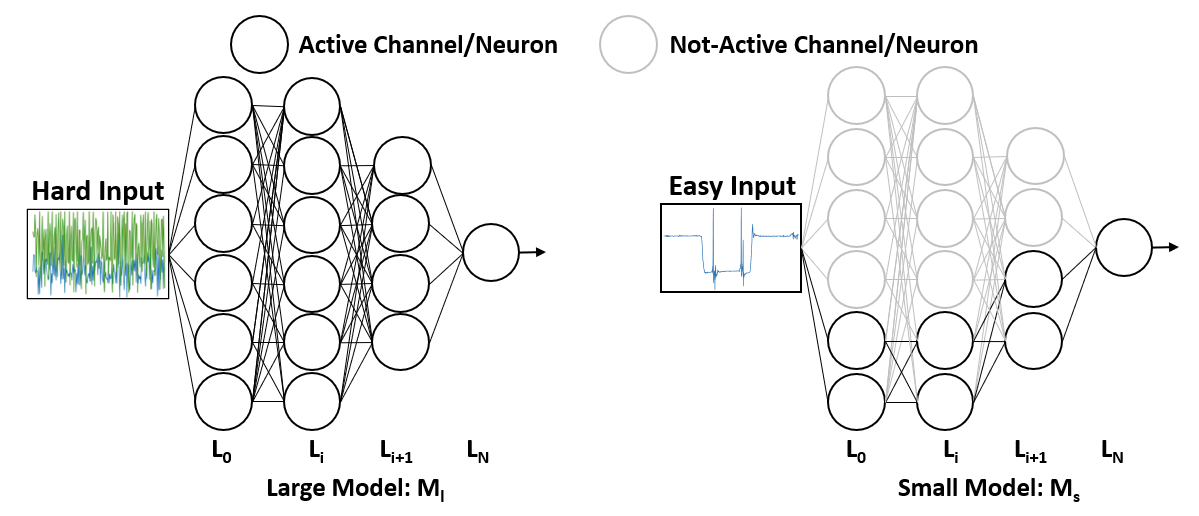}
    \caption{An overview of a variable-width input-adaptive neural network.}
    \label{fig:adaptive_slimmable}
\end{figure}

The adaptive technique that we employ in this work is an extension of the one proposed in~\cite{tann2016runtime} for 2D CNNs, and is based on ``variable-width'' networks. Figure~\ref{fig:adaptive_slimmable} shows a high level overview of the solution where, for simplicity, we display only a set of FC layers. 
The idea is to use a \textit{subset of the channels/features} of a large network to classify easy inputs. In this way, the small model ($M_s$) \textit{ideally} shares all of its parameters with the large one ($M_l$), resulting in no memory overhead with respect to a static version of the latter.
In practice, 
variable-width networks reach the best performance only if this zero-memory-overhead assumption is relaxed, as explained below.

In our work, besides extending the approach of~\cite{tann2016runtime} to 1D CNNs for HAR, we replace the original training procedure with a more recent and better performing one, proposed in~\cite{yu2018slimmable}. 
The procedure is based on repeating the forward and backward steps at all supported widths (i.e. number of ``active'' channels/features), for each batch of inputs. The corresponding gradients are accumulated, before performing a single optimization step. This is in contrast with the original incremental training of~\cite{tann2016runtime}, which trained different sub-networks sequentially to convergence, from the smallest to the largest, freezing the weights of the already-trained portions. 
Furthermore, we also inherit the idea of \textit{Switchable BatchNorm} from~\cite{yu2018slimmable}. This comes from the observation that, when using the above training procedure, BatchNorm layers do not work as expected. In fact, these layers normalize the output of Convolutional ones, by computing per-channel mean and standard deviation. When working with a reduced width, a subset of the channels is deactivated, causing those statistics to change. To cope with this problem, \cite{yu2018slimmable} proposed the use of private BatchNorm parameters for each supported width, showing that this technique yielded significant accuracy improvements, at the cost of a limited memory overhead. More details can be found in the original paper.
Importantly, the authors of~\cite{yu2018slimmable} did not focus on input-adaptive networks, but rather they trained so-called ``slimmable'' NNs, in which the number of active channels/features is controlled \textit{manually} by the user.
To our knowledge, we are the first to transfer the advanced training tricks of~\cite{yu2018slimmable} to an input-adaptive inference system.

Besides deciding how to implement the ``small'' and ``large'' models, another key element of the scheme in Figure~\ref{fig:adaptive_overview} is the implementation of the Adaptive Policy block.
A poorly designed policy, not able to correctly identify ``easy'' inputs, may render the entire system useless. In our work, we select a simple policy that has been demonstrated effective in literature~\cite{park2015big,tann2016runtime,JahierPagliari2018a,daghero2020energy}, which uses the difference between the largest two output scores produced by the little model, called \textit{Score Margin} (SM), 
as a measure of its classification confidence. Thus, the global output of the adaptive system is determined as follows:
\begin{equation}
    \hat{y} =
    \begin{cases}
    \hat{y}_s\textrm{ if }SM=max(\hat{y}_s)-2nd\_max(\hat{y_s}) > T_h\\
    \hat{y_l}\textrm{ otherwise}
    \end{cases}
\end{equation}
where $\hat{y}_s$ and $\hat{y}_l$ are the outputs of the small and large model, respectively, and
$T_h$ is a user-defined threshold. 
The rationale of SM-based policies is that if $M_s$ has produced a large output probability for \textit{one class only}, then it is ``confident'' that such class is the correct one.
$T_h$ defines what is considered ``high-confidence'', in turn determining $\mathbb{P}_e$ in (\ref{eq:exp_cost}). Therefore, different operating modes can be obtained simply by tuning a single scalar ($T_h$) at runtime.

An issue linked with the implementation of adaptive systems is how to choose suitable static architecture(s) to use as starting point. In the specific case of variable-width networks, this corresponds to selecting the hyper-parameters of the largest model $M_l$, as well as the number of channels/features deactivated to form the small model $M_s$.
To our knowledge, these design choices have not been investigated extensively in literature, and in most cases, they are assumed as given.
Having performed an extensive architecture search for each target dataset (see Section~\ref{sec:grid_search}), we have at our disposal a rich Pareto front of static models to choose from.
Therefore, we derived a systematic way to select promising architectures to convert into adaptive systems. 

We observe that taking as $M_l$ the \textit{most accurate} CNN overall from our design space is not ideal. In fact, as shown by our results of Section~\ref{sec:results}, large networks provide diminishing returns in terms of accuracy versus inference cost: they often achieve a little accuracy gain in exchange for an explosion in inference cycles (and hence latency and energy consumption). Consequently, a reduced-width version of the most accurate network would still be very complex, and would achieve a similar accuracy even on hard inputs, nullifying the principle of input-adaptive inference. On the other hand, selecting a too small/inaccurate $M_l$ would make the corresponding sub-network $M_s$ unable to correctly estimate output scores, rendering the SM policy inaccurate. Accordingly, we instead select $M_l$ by computing the following ``gain'' metric $G_i$ for each network in our Pareto frontier:
\begin{equation}\label{eq:gain}
    G_{i}=\frac{P(M_i) - P(M_{i-1})}{C(M_i) - C(M_{i-1})}
\end{equation}
where $P(M_i)$ is the classification score (accuracy or other metric) of the $i$-th model, and $C(M_i)$ is its inference cost (latency or energy). The index $i$ refers to the Pareto-optimal models ordered by increasing accuracy. We then select as $M_l$ the model with the largest $G_i$, among those with an accuracy drop $< 5\%$ with respect to the most accurate network in the set.
Lastly, we derive the small model $M_s$ by activating either the first 25\% or the first 50\% channels/features in each layer of the network, and selecting, between the two, the version that yields the best accuracy versus cost trade-off, with the expected cost formulation of (\ref{eq:exp_cost}).

\section{Experimental Results}\label{sec:results}

\subsection{Setup}

We train all our CNNs using Python 3.8 and the PyTorch deep learning framework~\cite{pytorch}, following a common scheme for all datasets.
Specifically, we employ the Adam optimizer, with a Learning Rate (LR) scheduler that multiplies the LR by a factor 0.1 when the training loss does not decrease for 3 consecutive epochs. We also perform early-stopping after 5 epochs of stale training.
For all datasets, we use the categorical cross entropy loss as $\mathcal{L}_{task}$, with class-dependent loss weights equal to the inverse of the training set class frequencies.
For the UniMiB-SHAR, WISDM and WALK datasets, we use a batch size of 32. As mentioned in Section~\ref{sec:datasets}, the train/test splits are the same proposed in~\cite{comparison_wisdm_unimib_dnn}, \cite{wisdm} and \cite{daghero2021ultracompact} respectively, while the initial LR is 0.001 for the first two datasets, and 0.01 for WISDM. For UniMiB-SHAR, we also exploit weight decay with a regularization strength of $10^{-4}$. Lastly, for the UCI HAPT dataset, we set the initial LR to 0.01, the weight decay strength to $10^{-4}$ and the batch size to 128, keeping the train/test split of~\cite{hapt}.
The RFs implemented as baselines for comparison are trained using the Scikit-Learn package~\cite{scikit-learn} with the same configuration described in~\cite{daghero2021ultracompact}, whereas when comparing with state-of-the-art deep learning models, we take the accuracy and complexity results directly from the original papers.

We report classification results using various metrics depending on the dataset, in order to compare fairly with previous literature. Specifically, besides the standard classification \textit{accuracy}, we consider two additional metrics, which are more relevant for class-imbalanced datasets. Namely, we compute the \textit{balanced accuracy} (B. Accuracy), i.e., the arithmetic mean of recall over all classes, and the \textit{F1-Score} (F1), i.e., the harmonic mean of precision and recall.
All classification results reported in the following sections refer to the test sets of each dataset.

Trained and quantized CNNs are converted to C code using the inference back-end library described in~\cite{Burrello2021b} and the sub-byte kernels of~\cite{bruschi2020enabling} for $>$ 1-bit precision, both adapted to work on the single-core PULPissimo. For binary layers, and for the RF implementations used as comparison, we resort to the implementations presented in~\cite{daghero2021ultracompact}.

\subsection{Memory Occupation}

\begin{figure*}[ht]
\centering
\includegraphics[width=.9\textwidth]{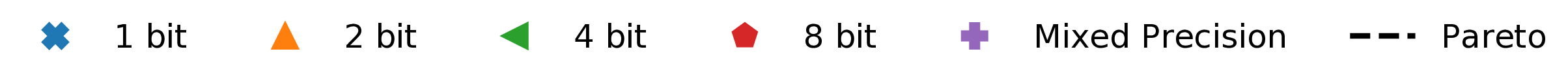}
\begin{subfigure}{0.249\linewidth}%
\includegraphics[width=\linewidth]{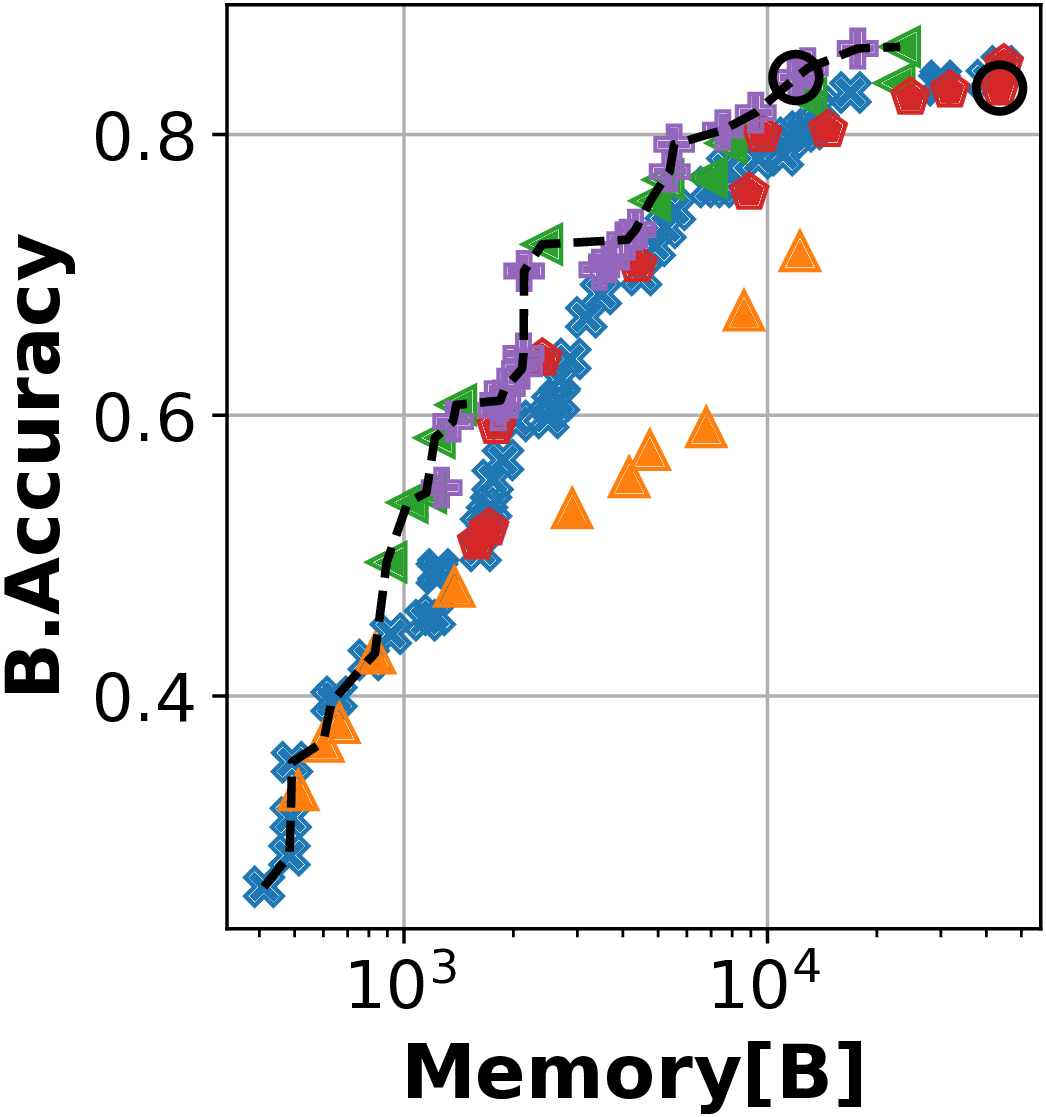}%
\subcaption{UniMiB-SHAR}\label{fig:unimib_memory}%
\end{subfigure}%
\begin{subfigure}{0.249\linewidth}%
\includegraphics[width=\linewidth]{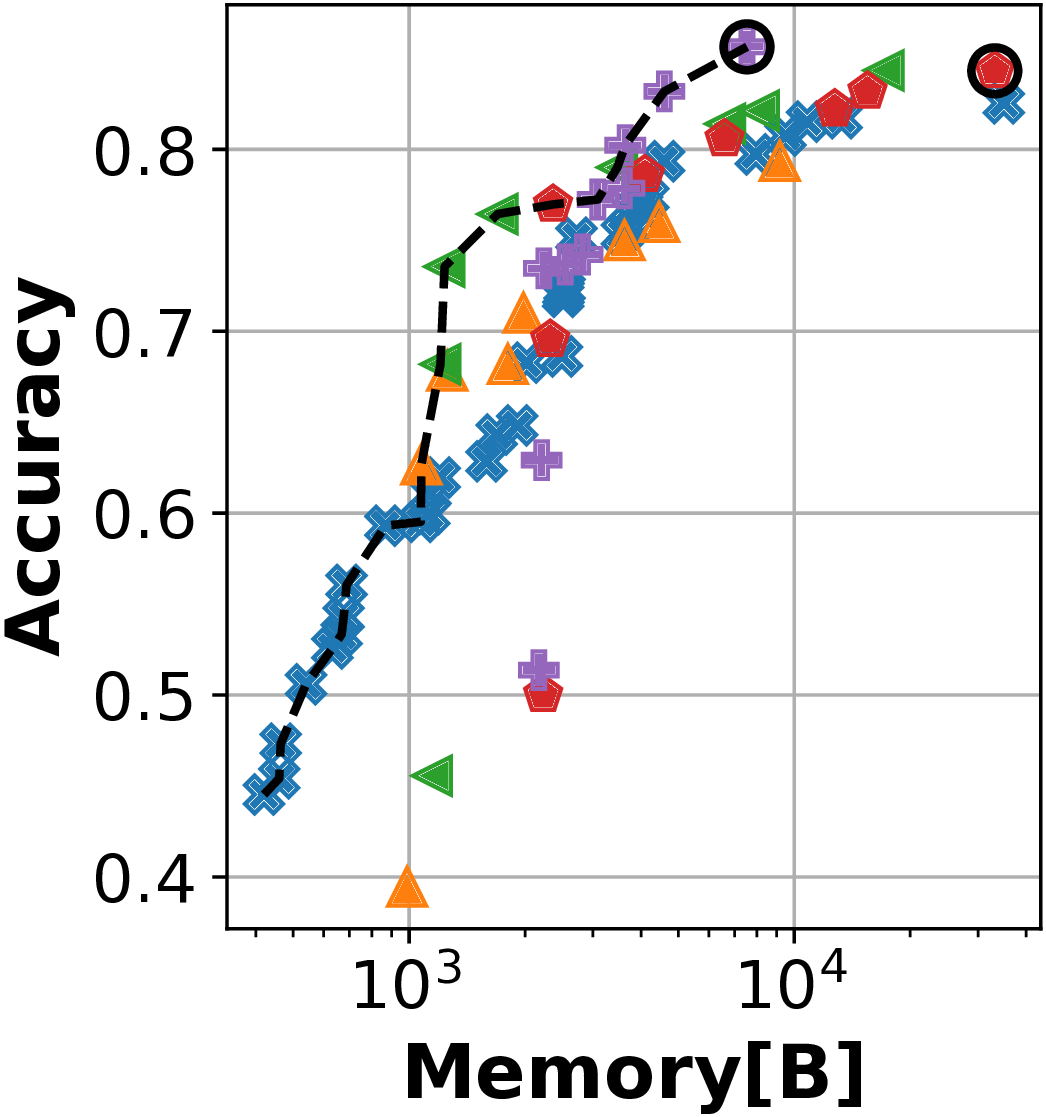}%
\subcaption{UCI HAPT}\label{fig:hapt_memory}%
\end{subfigure}%
\begin{subfigure}{0.249\linewidth}%
\includegraphics[width=\linewidth]{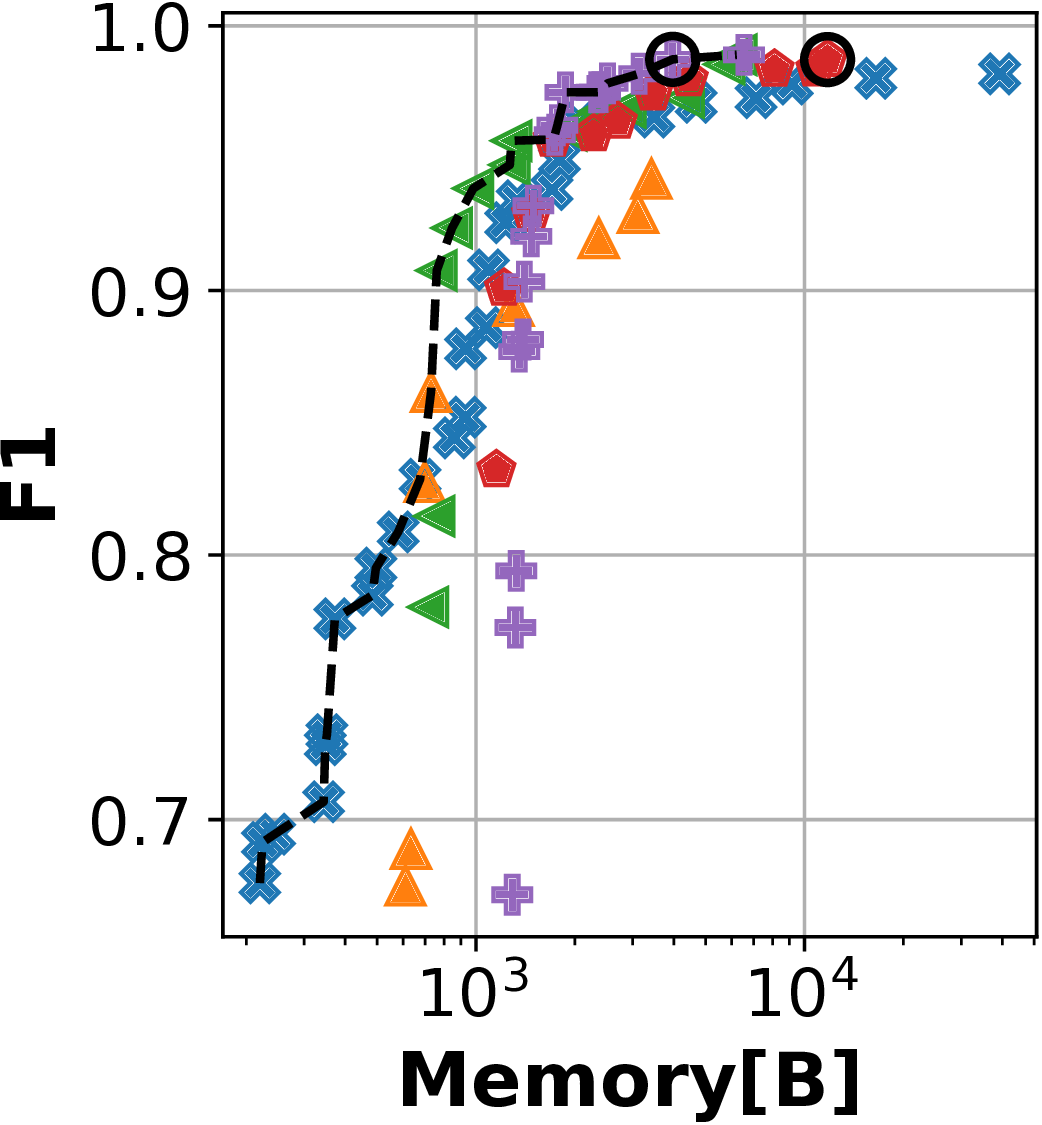}%
\subcaption{WISDM}\label{fig:wisdm_memory}%
\end{subfigure}%
\begin{subfigure}{0.249\linewidth}%
\includegraphics[width=\linewidth]{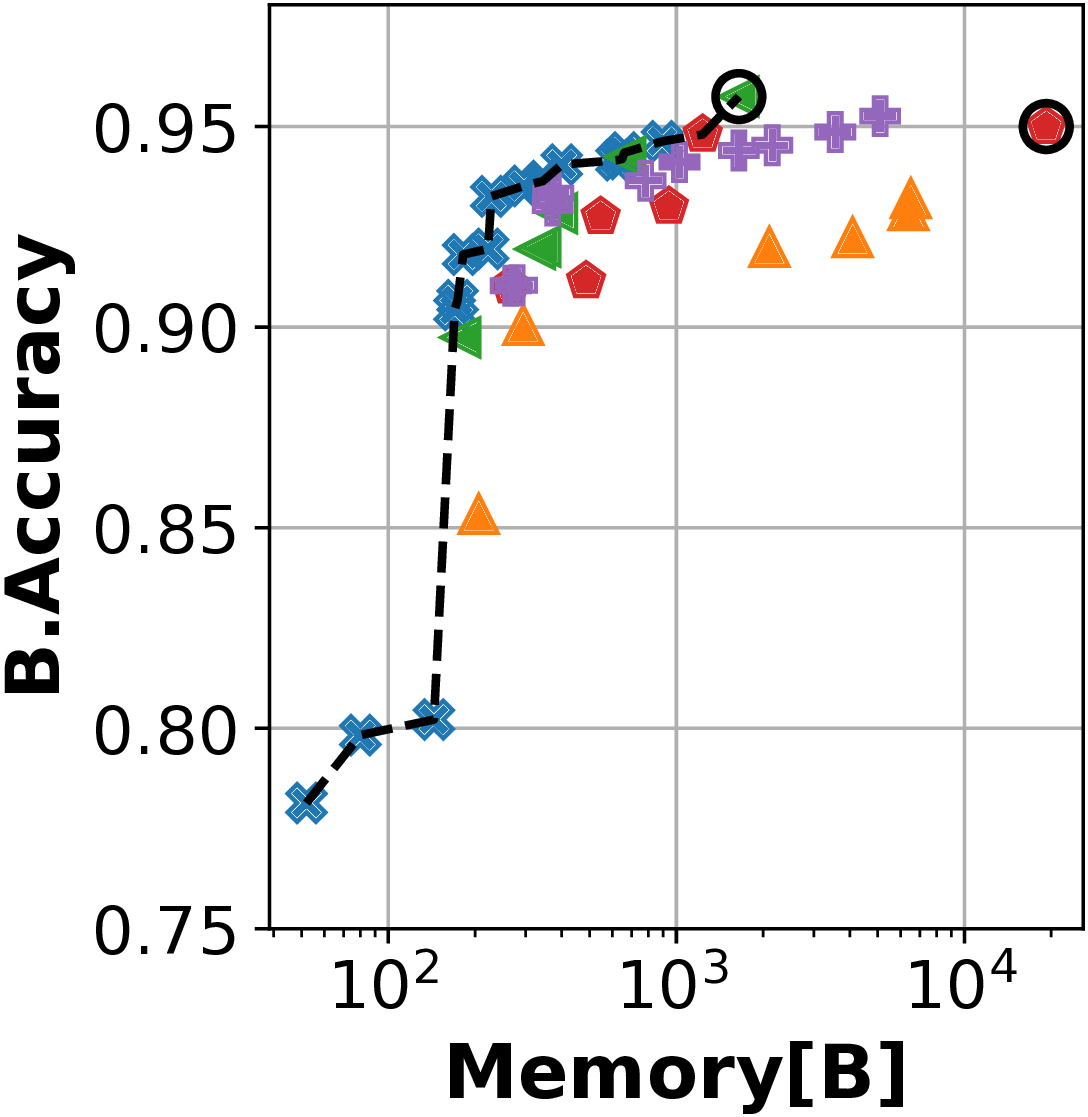}%
\subcaption{WALK}\label{fig:walk_memory}
\end{subfigure}%
\caption{Classification score versus memory occupation Pareto fronts, for different quantization formats. The black dashed line shows the global Pareto front.}\label{fig:memory_optimization_results}
\end{figure*}

Figure~\ref{fig:memory_optimization_results} shows the results of our memory optimization flow. Specifically, the graphs show the Pareto fronts in terms of classification score (Accuracy, B. Accuracy, or F1) versus memory occupation obtained combining architectural exploration with grid search (Section~\ref{sec:grid_search}) and mixed-precision search (Section~\ref{sec:memory_opt}). Each color refers to one type of quantization, and points correspond to Pareto-optimal CNN architectures (with different layer hyper-parameters) that use such quantization. The overall Pareto front is highlighted by a dashed black line. All graphs have logarithmic x axes.

The figure shows similar trends for three of the four considered benchmarks. Results on UniMiB-SHAR (Figure~\ref{fig:unimib_memory}), i.e., the most complex dataset in terms of number of classes, show that 1-bit, 4-bit and mixed-precision CNNs obtain Pareto-optimal performance, in the low ($<$ 45\%), mid, and high ($>$ 75\%) balanced accuracy regimes, respectively.
At best, mixed-precision networks reduce the memory occupation of up to 72\% with respect to 8-bit ones, with no B. Accuracy drop. The 8-bit and mixed precision points corresponding to this maximum memory reduction are highlighted in the figure with two black circles.
Moreover, due to the well known regularizing effect of quantization~\cite{Jacob2018}, the best balanced accuracy achieved by sub-byte CNNs is even higher than the one obtained by 8-bit ones (+1.23\%).

Similar considerations apply to the results on UCI HAPT, the second most complex of our benchmarks (Figure~\ref{fig:hapt_memory}). 4-bit networks are again on the global Pareto curve for architectures with intermediate size, whereas mixed-precision CNNs achieve the best trade-off for high accuracy values ($> 80\%$), obtaining them with far less parameters than any fixed-precision solution.
Precisely, our best mixed-precision CNN obtains an accuracy of 85.63\%, i.e., +1.3\% with respect to the best 8-bit quantized network, with a memory reduction of 77\%.
At the other end of the Pareto curve (accuracy $<60\%$) full 1-bit quantization (binarization) also configures as an interesting alternative, generating extremely small yet still accurate models (considering that this dataset has 12 classes).

The WISDM dataset shows again a similar trend (Figure~\ref{fig:wisdm_memory}), despite supporting a lower number of classes than the previous two and using yet another classification metric. In this case, however, all quantization bit-widths above 2-bit achieve very similar trade-offs. Mixed-precision CNNs are able to reduce the memory occupation by up 66\% with respect to 8-bit networks for the same F1-score.

WALK is the only dataset that exhibits a different trend (Figure~\ref{fig:walk_memory}), with BNNs occupying most of the global Pareto-curve. This is mainly due to the simpler binary classification task, which calls for more compact models (see the different range of the x axis). A fixed-precision 4-bit network reaches the highest balanced accuracy of  95.74\%, while saving 91\% of the memory when compared to the most accurate 8-bit network.
\begin{figure*}[ht]
\centering%
\includegraphics[width=.38\textwidth]{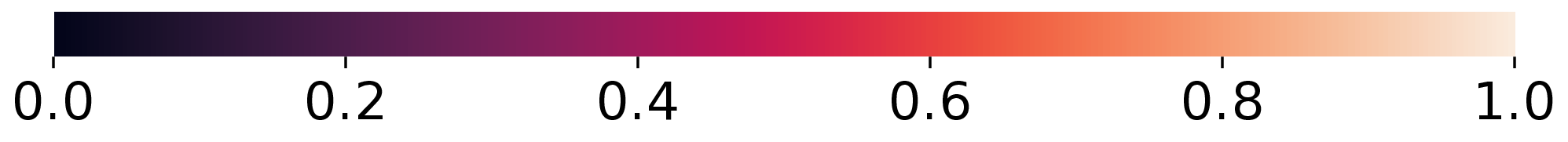}

\begin{subfigure}{0.18\linewidth}%
\includegraphics[width=\linewidth]{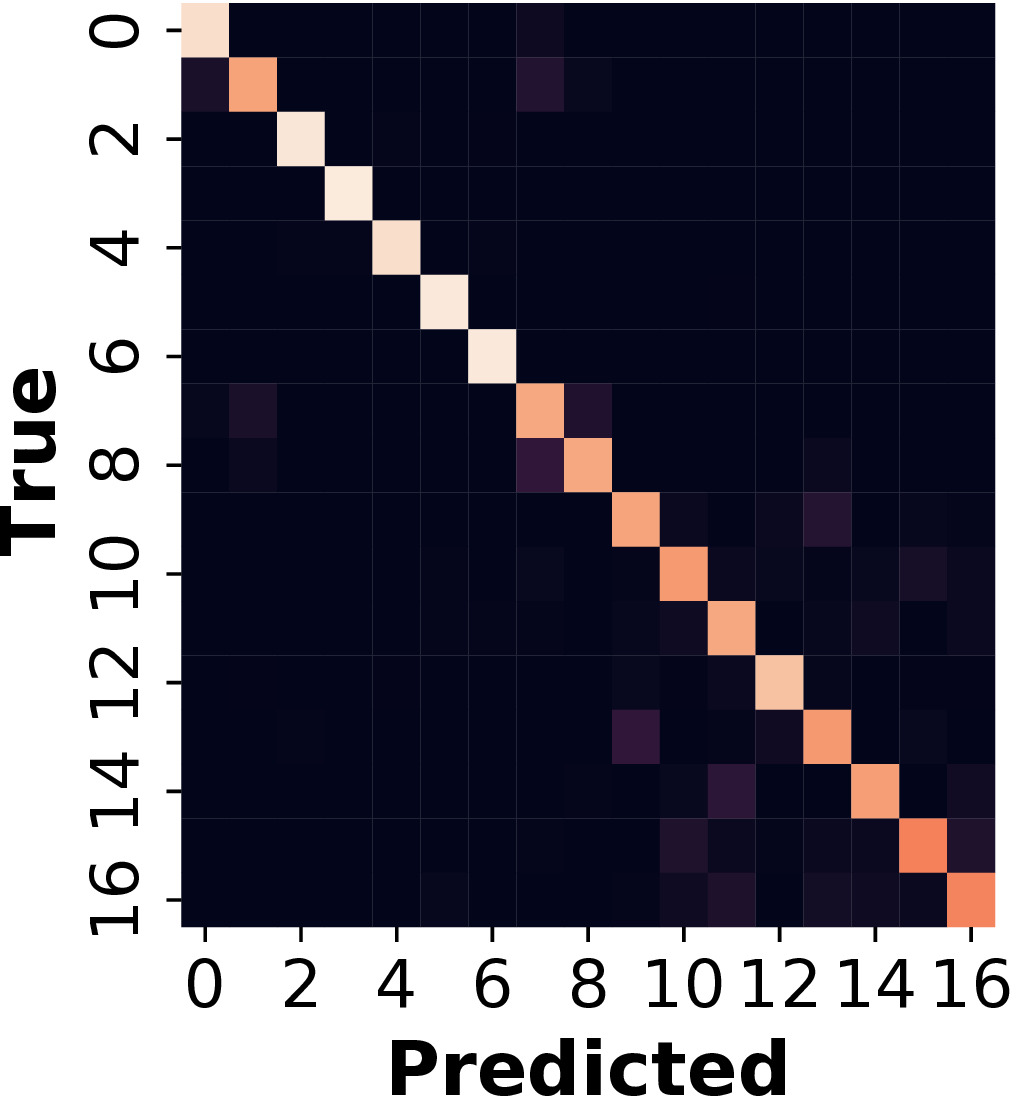}%
\subcaption{UniMiB-SHAR}%
\label{fig:unimib_cm}%
\end{subfigure}%
\begin{subfigure}{0.18\linewidth}%
\includegraphics[width=\linewidth]{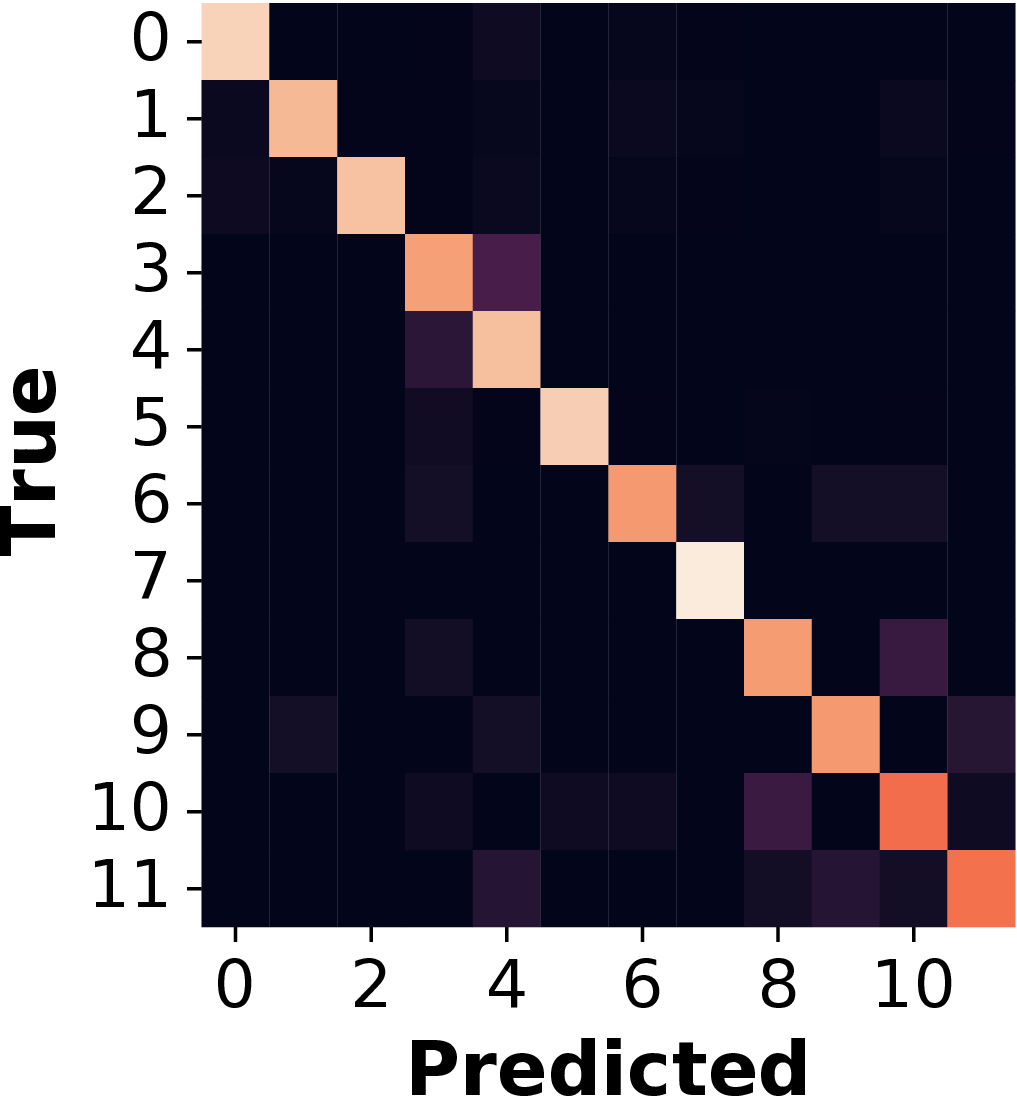}%
\subcaption{UCI HAPT}%
\label{fig:hapt_bn}%
\end{subfigure}%
\begin{subfigure}{0.18\linewidth}%
\includegraphics[width=\linewidth]{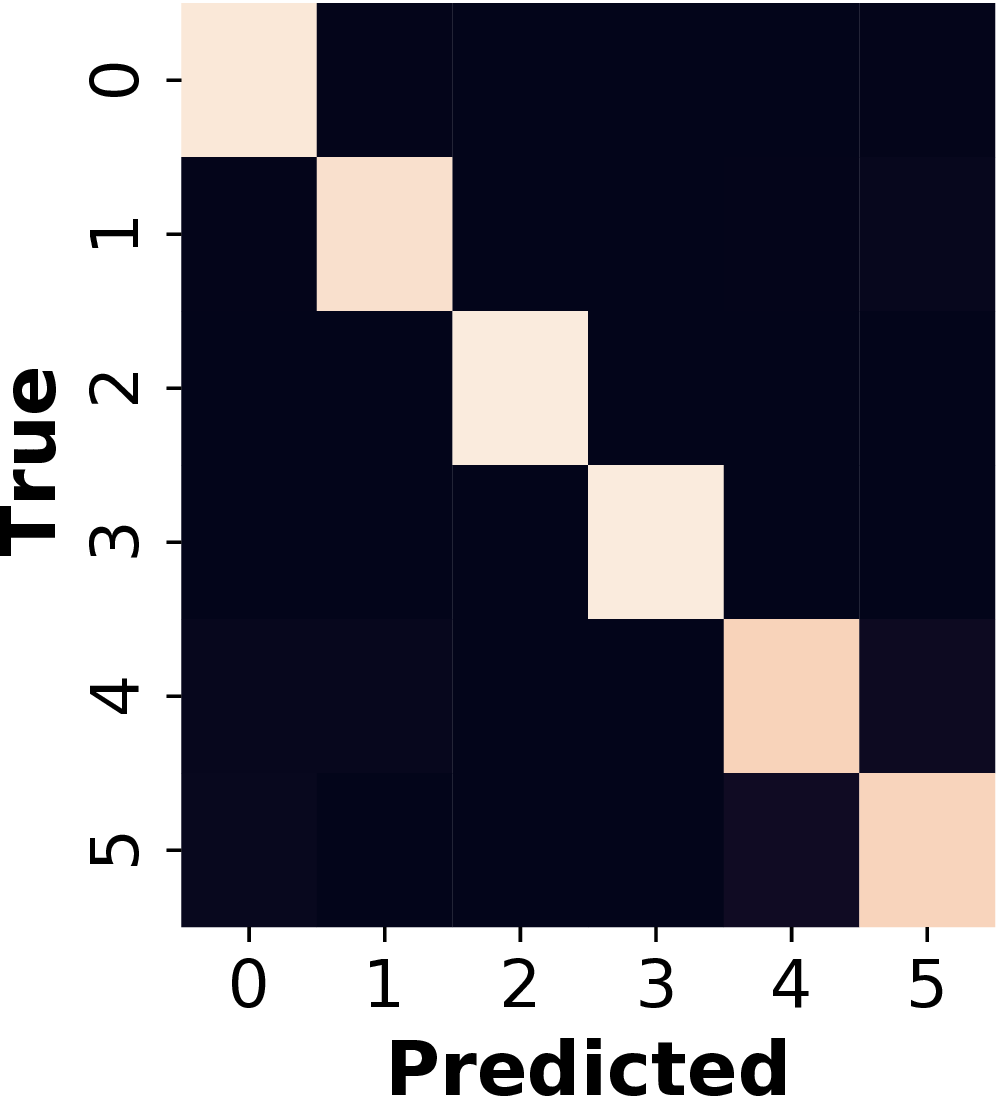}%
\subcaption{WISDM}%
\label{fig:wisdm_cm}%
\end{subfigure}%
\begin{subfigure}{0.18\linewidth}%
\includegraphics[width=\linewidth]{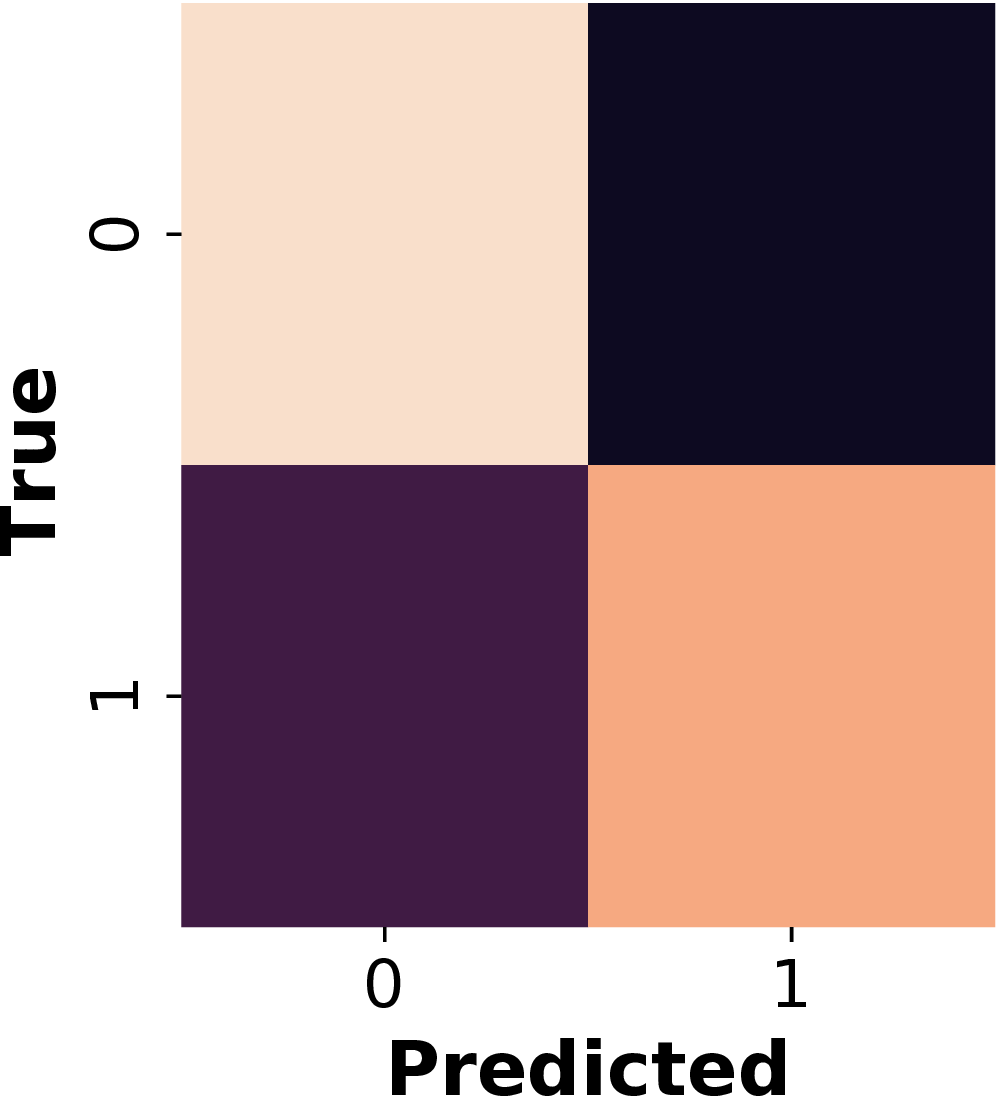}%
\subcaption{WALK}%
\label{fig:walk_cm}
\end{subfigure}%
\caption{Confusion Matrices of the most accurate models for each dataset, normalized by the true labels.}%
\label{fig:cm_comparison}%
\end{figure*}%
Figure~\ref{fig:cm_comparison} shows the confusion matrix of the most accurate model for each dataset as a heat map, confirming that the networks found during the proposed search, despite their small size and low-precision quantization, are able to effectively differentiate among classes, handling correctly the imbalance of the considered HAR datasets. 

Overall, these results show that using sub-byte quantization, both at fixed- and mixed-precision, permits significant memory savings for HAR tasks with no classification score drop. In many cases, sub-byte networks also achieve a higher maximum score compared to 8-bit ones, thanks to the reduced over-fitting induced by low precision data representation. 

Since we target the execution of HAR model on MCUs, our exploration only considers integer quantization formats up to 8-bit, as detailed in Section~\ref{sec:methods}. This is because many low-power MCUs, including the one used in our deployment experiments, do not have a hardware Floating-Point Unit (FPU), hence they cannot perform float operations efficiently. Nonetheless, we still trained the most accurate CNN found for each dataset  without QAT, as an ablation study. This experiment confirmed that quantization does not cause scores drops, and actually results in a slight increase in performance, once again thanks to its well-known regularization effect. The gains range from a minimum of +0.1\% on UCI HAPT to a maximum of +4\% on WALK.

\subsection{Execution Cycles}

\begin{figure*}[ht]
\centering
\includegraphics[width=.9\textwidth]{figures/results/mixed/base/legenda}
\begin{subfigure}{0.24\linewidth}%
\includegraphics[width=\linewidth]{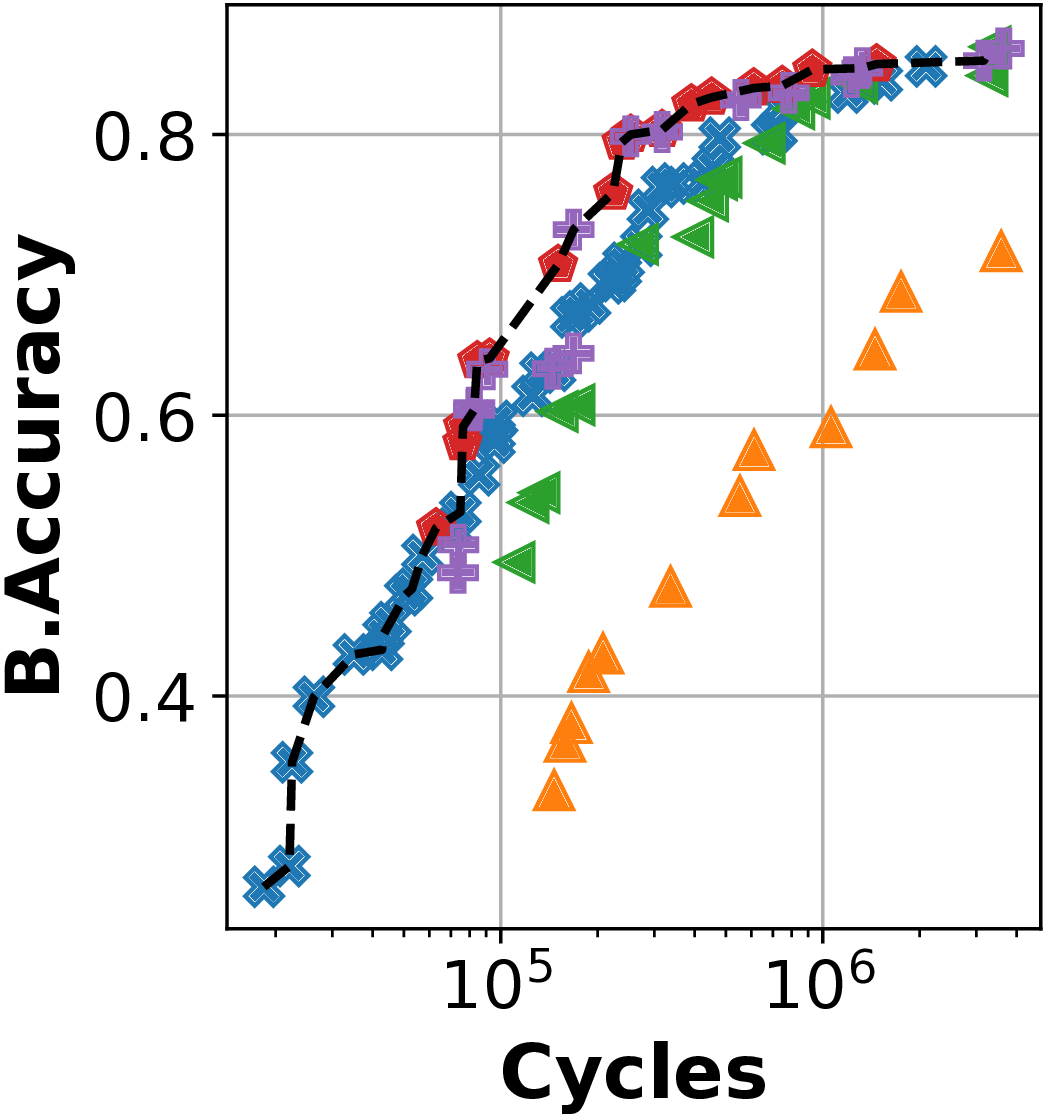}%
\subcaption{UniMiB-SHAR}\label{fig:unimib_cycles}%
\end{subfigure}%
\begin{subfigure}{0.24\linewidth}%
\includegraphics[width=\linewidth]{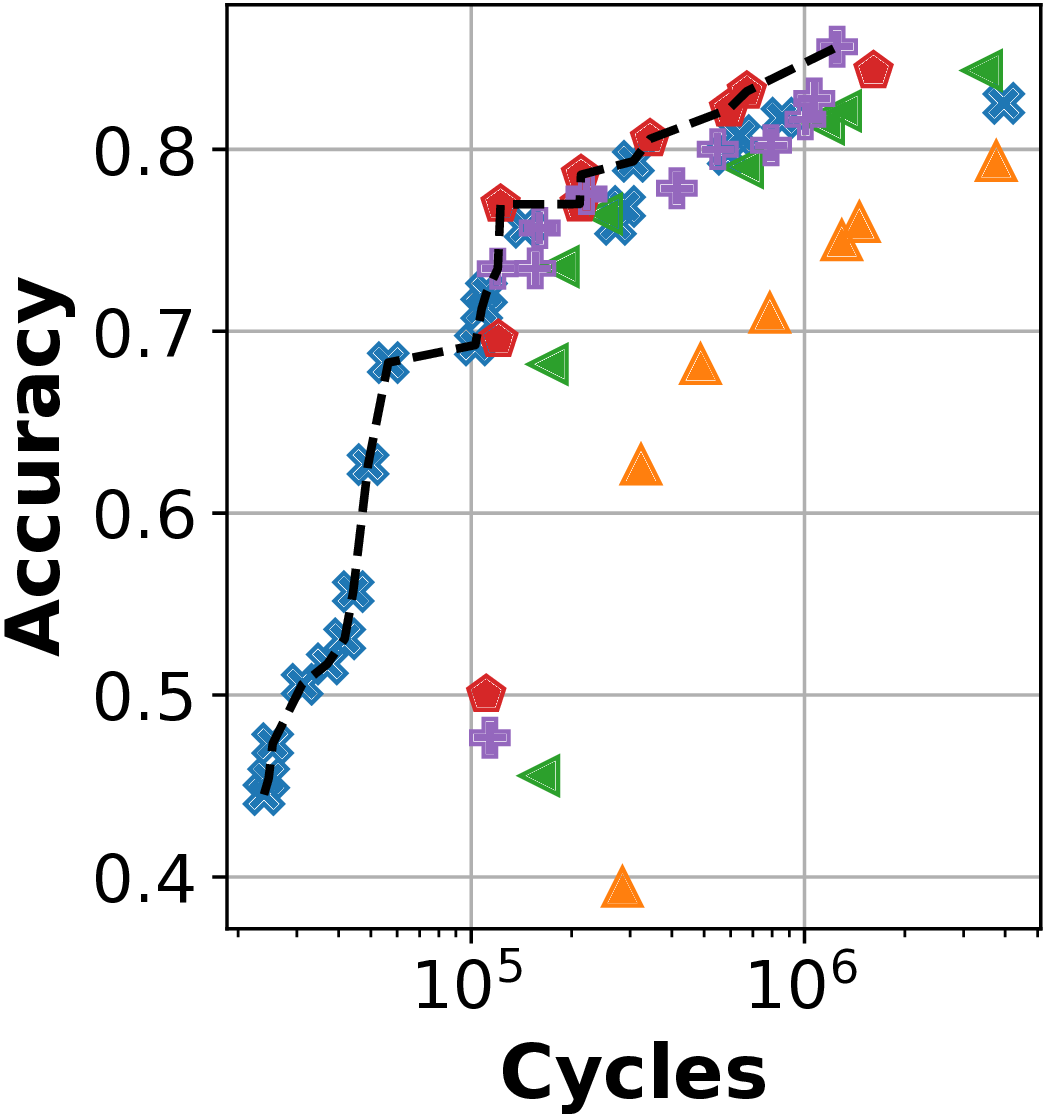}%
\subcaption{UCI HAPT}\label{fig:hapt_cycles}%
\end{subfigure}%
\begin{subfigure}{0.24\linewidth}%
\includegraphics[width=\linewidth]{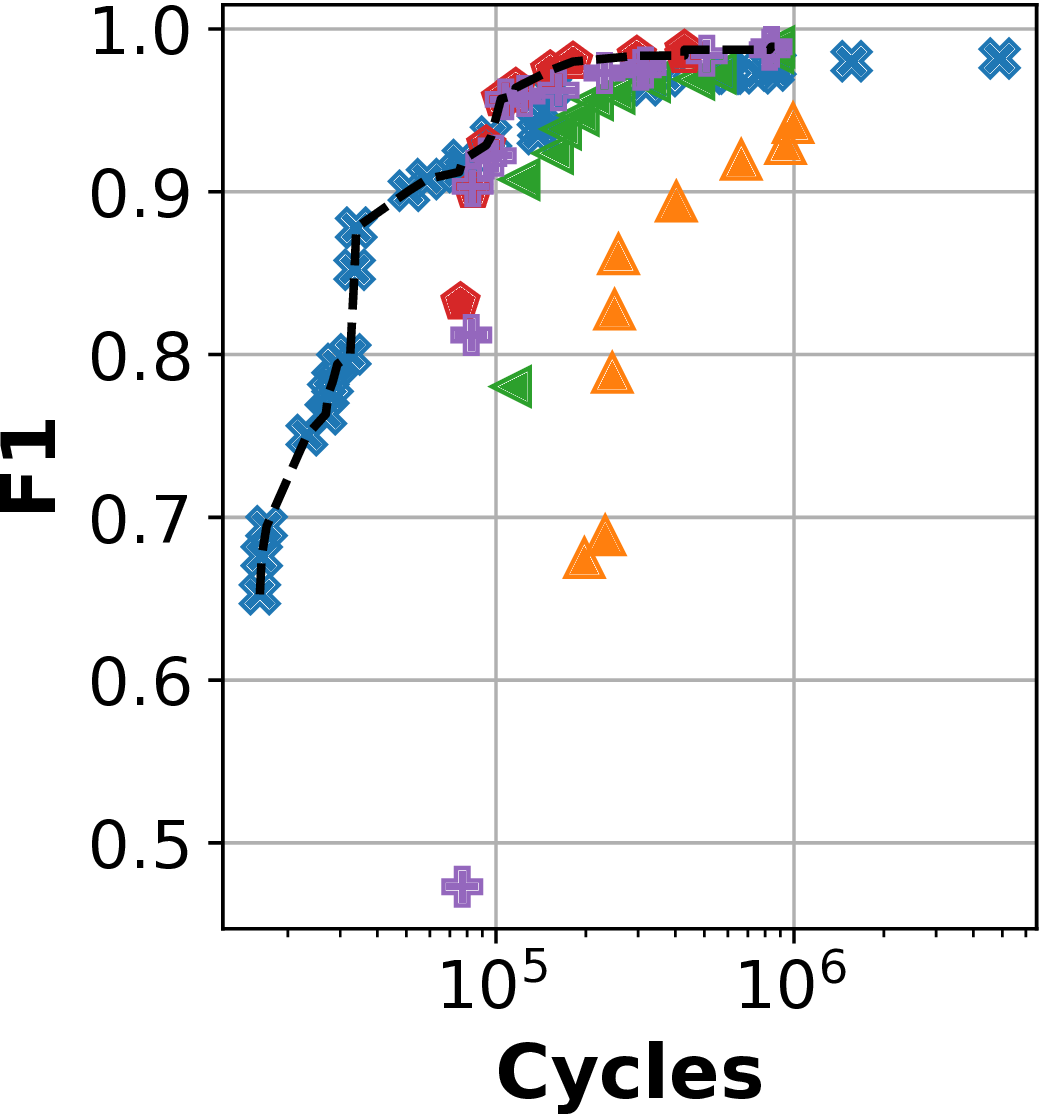}%
\subcaption{WISDM}\label{fig:wisdm_cycles}%
\end{subfigure}%
\begin{subfigure}{0.24\linewidth}%
\includegraphics[width=\linewidth]{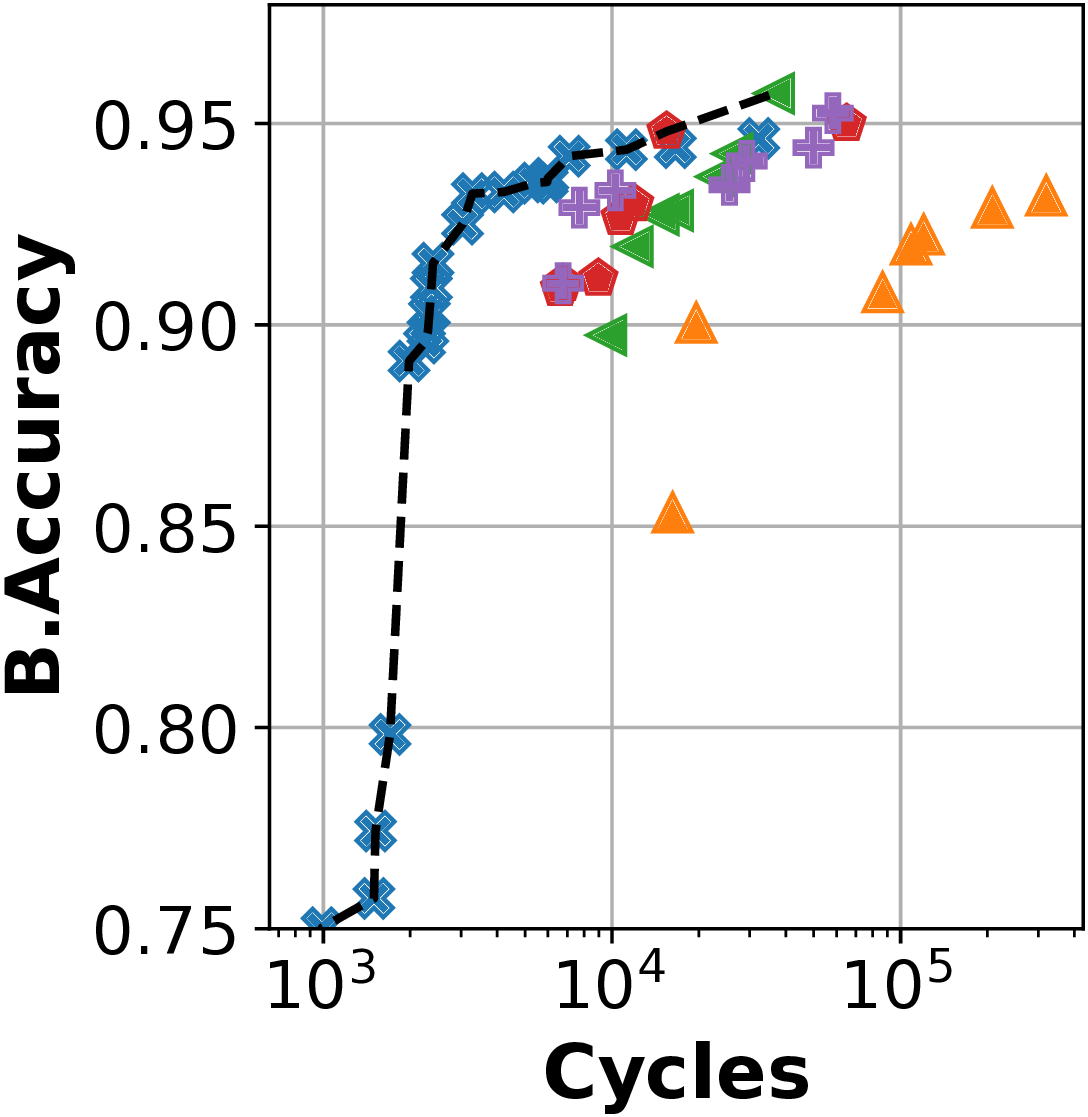}%
\subcaption{WALK}\label{fig:walk_cycles}%
\end{subfigure}%
\caption{Classification score versus inference cycles Pareto fronts, for different quantization formats. The black dashed line shows the global Pareto front.}\label{fig:cycles_optimization_results}
\end{figure*}

Figure~\ref{fig:cycles_optimization_results} shows the results obtained by our static 1D CNNs in terms of activity recognition score versus number of inference clock cycles on the target MCU. The number of cycles is a direct proxy of the total inference latency and energy consumption. Points and colors have the same meaning of Figure~\ref{fig:memory_optimization_results}, but the Pareto-optimal CNNs are in general different, since an architecture that achieves a good accuracy versus memory trade-off isn't necessarily also optimal in terms of accuracy versus inference cycles.

The figure shows that, when considering the number of inference clock cycles, fixed-precision sub-byte CNNs are no longer Pareto-optimal, with the exception of BNNs (for low classification scores). Instead, the Pareto frontier in the high scores regime is entirely occupied by 8-bit and mixed-precision networks. This is because, as explained in Section~\ref{sec:memory_opt}, sub-byte quantization does not have a beneficial effect on execution speed on general purpose MCUs, which do not natively support operands with less than 8-bit. On the contrary, the implementations of sub-byte layers used in this work~\cite{bruschi2020enabling} are based on un-packing data read from memory onto 1 byte with bit-extension operations, performing multiply-and-accumulate operations on 8-bit, and finally re-packing the results onto the appropriate bit-width. This slows down the execution, moving sub-byte networks away from the Pareto frontier. These considerations do not apply to BNNs, which thanks to the efficient implementation of~\cite{daghero2021ultracompact} and to the replacement of arithmetic operations with bit-wise logic ones, are very effective also in reducing execution cycles. Lastly, mixed-precision CNNs still achieve good results because, by mixing 8-bit and sub-byte layers, they incur a lower cycles overhead due to un-packing and packing operations.

\begin{table}[ht]
\footnotesize
\begin{tabular}{l|ccc}
\hline
\textbf{Dataset}     & \textbf{Score Range [\%]} & \textbf{Memory Range [kB]} & \textbf{Cycles Range $[\cdot 10^3]$} \\\hline
\textit{UNIMIB-SHAR} & 26.24:86.24 [BAcc.]    & 0.41:23.17               & 18.4:3316\\
\textit{UCI HAPT}    & 44.53:85.63 [Acc.]    & 0.42:7.54                & 24:1253\\
\textit{WISDM}       & 67.6:98.9 [F1]      & 0.22:6.22                & 16:859\\
\textit{WALK}        & 78.13:95.74 [BAcc.]    & 0.05:1.65               & 0.9:36.4\\\hline
\end{tabular}\caption{Summary of the characteristics of Pareto-optimal models for each dataset.}\label{tab:deployment_stats}
\end{table}

Table~\ref{tab:deployment_stats} summarizes the results of our static models exploration. Specifically, it shows the ranges of classification score, memory occupation and inference cycles of the models lying on the Pareto curve of each dataset. Globally, Pareto CNNs span almost up to 2 orders of magnitude in memory and cycles, and up to $\pm 60\%$ in terms of classification metrics, showing the breadth of our exploration.

\subsection{Adaptive Inference}

\begin{figure*}[ht]
\centering
\includegraphics[width=.5\textwidth]{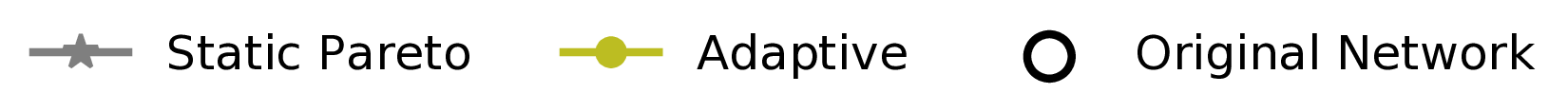}
\begin{subfigure}{0.24\linewidth}%
\includegraphics[width=\linewidth]{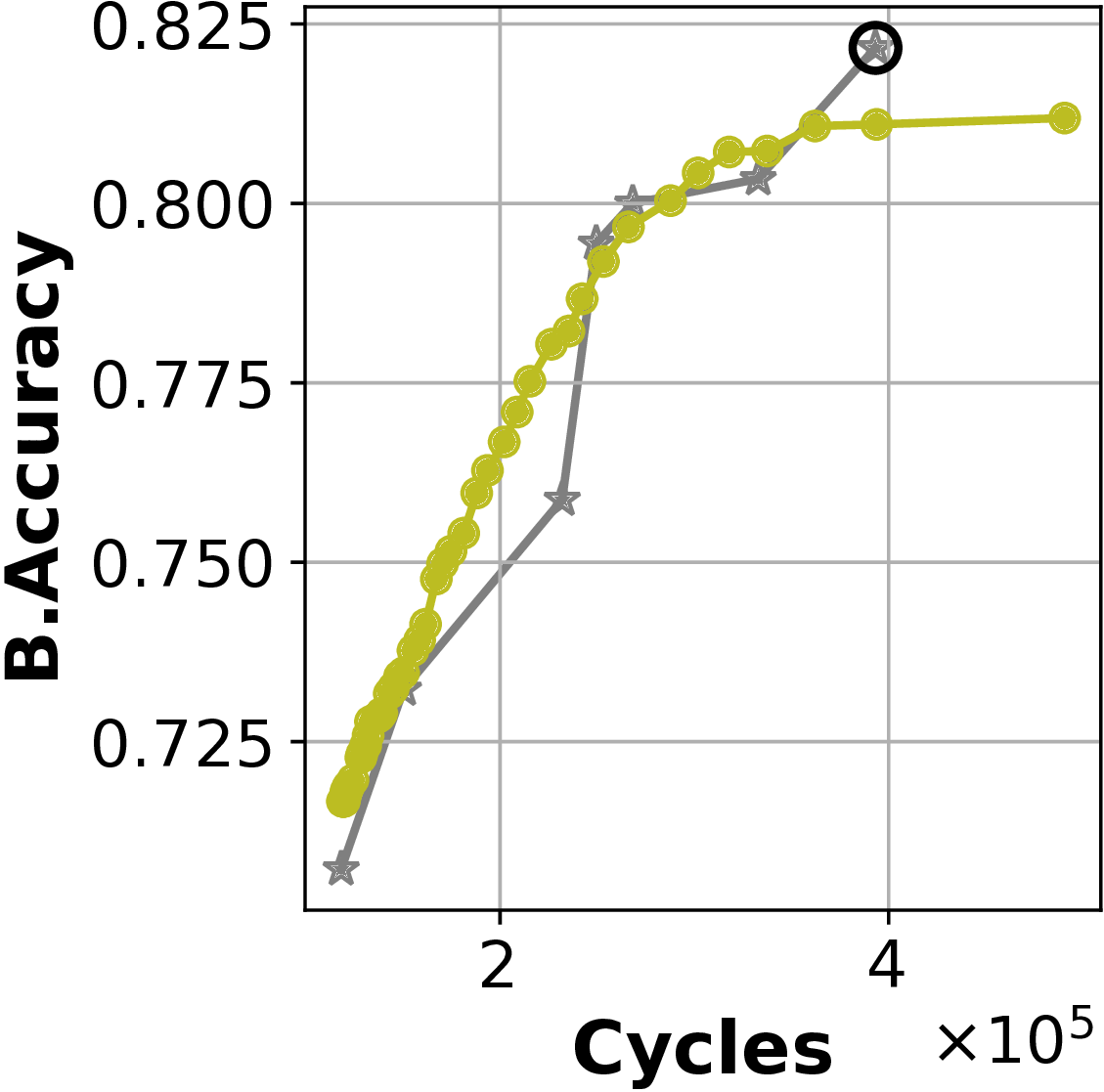}%
\subcaption{UniMiB-SHAR}\label{fig:unimib_adaptive}%
\end{subfigure}%
\begin{subfigure}{0.24\linewidth}%
\includegraphics[width=\linewidth]{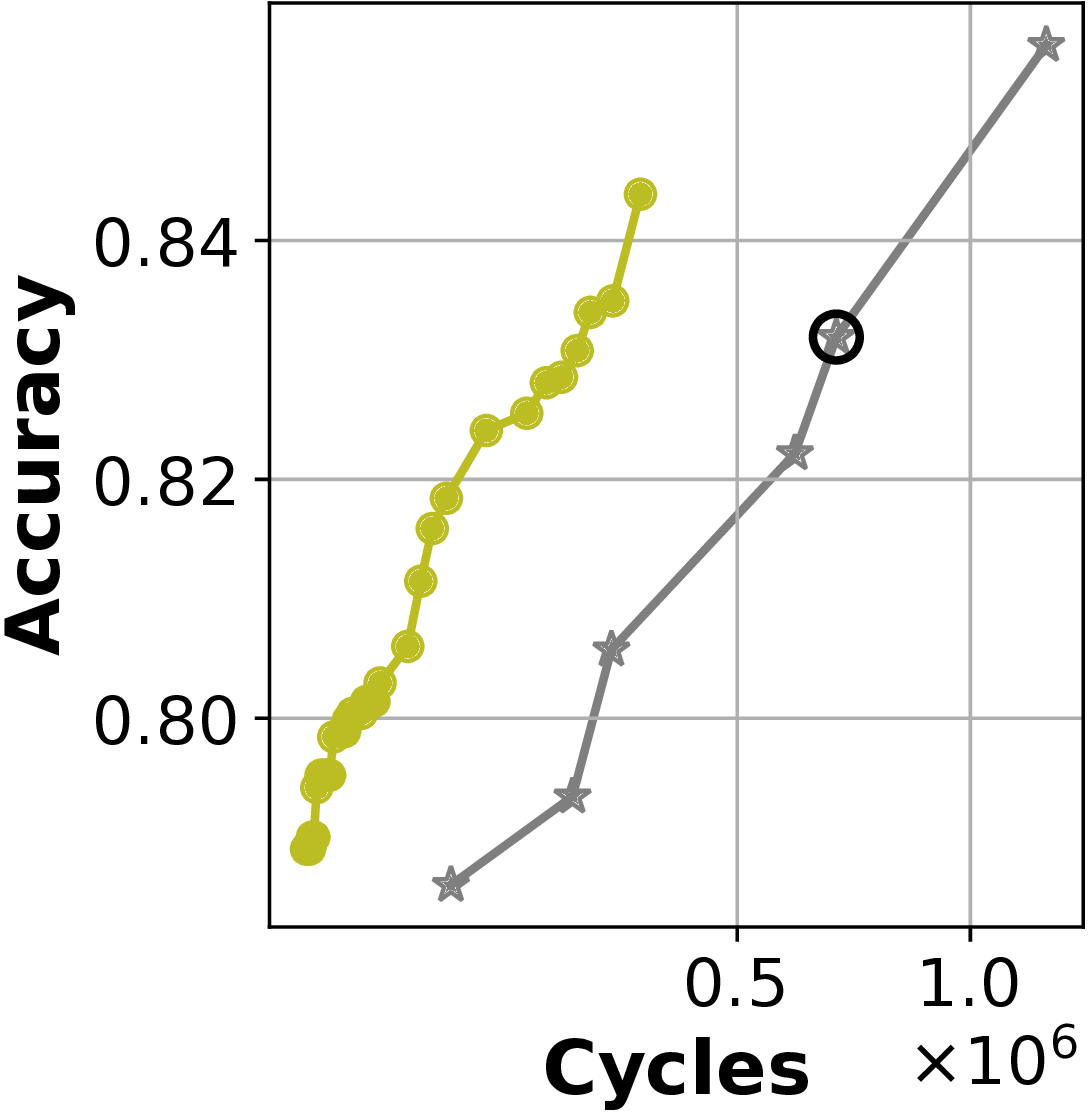}%
\label{fig:hapt_adaptive}\subcaption{UCI HAPT}%
\end{subfigure}%
\begin{subfigure}{0.24\linewidth}%
\includegraphics[width=\linewidth]{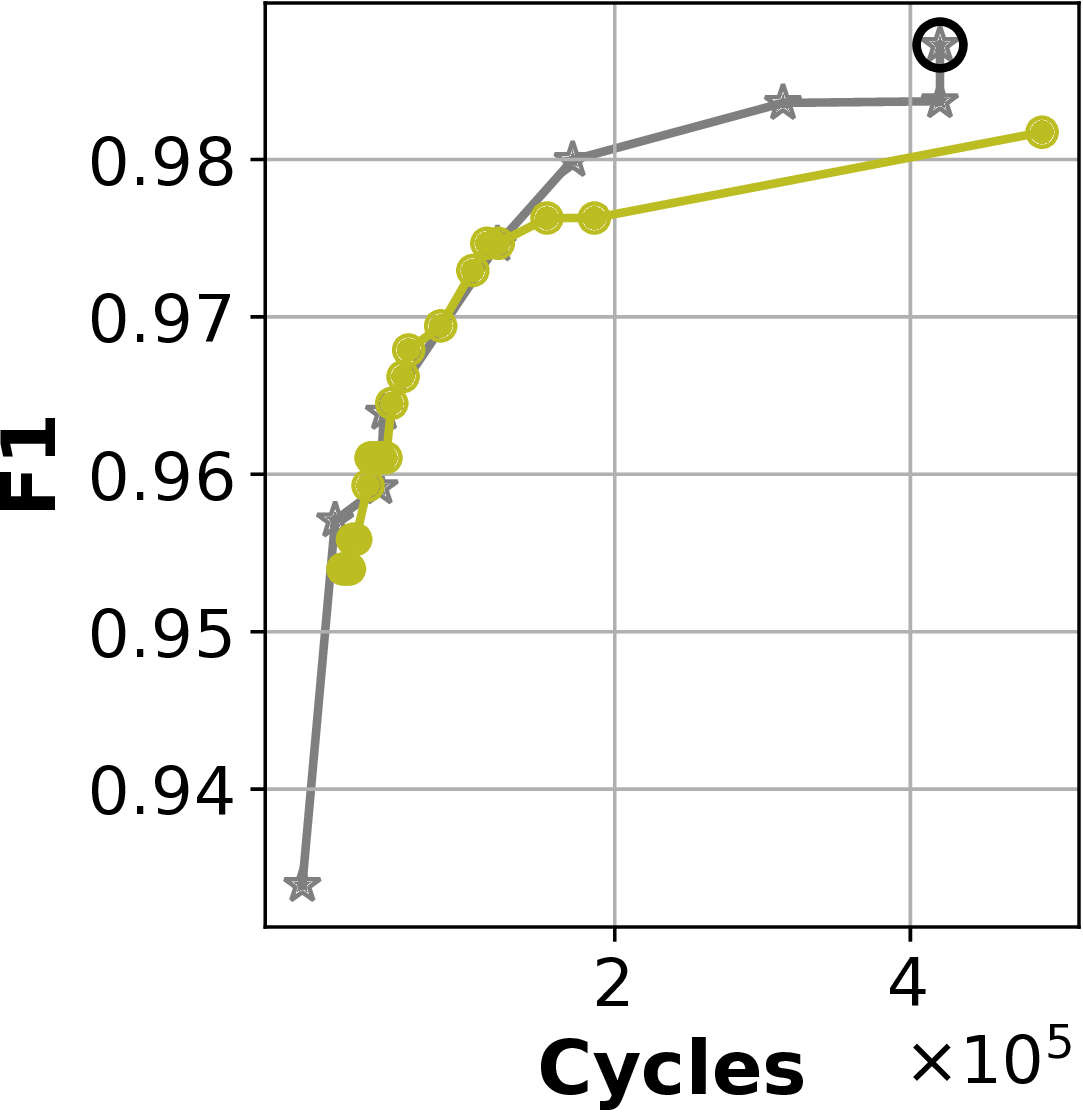}%
\subcaption{WISDM}\label{fig:wisdm_adaptive}%
\end{subfigure}%
\begin{subfigure}{0.24\linewidth}%
\includegraphics[width=\linewidth]{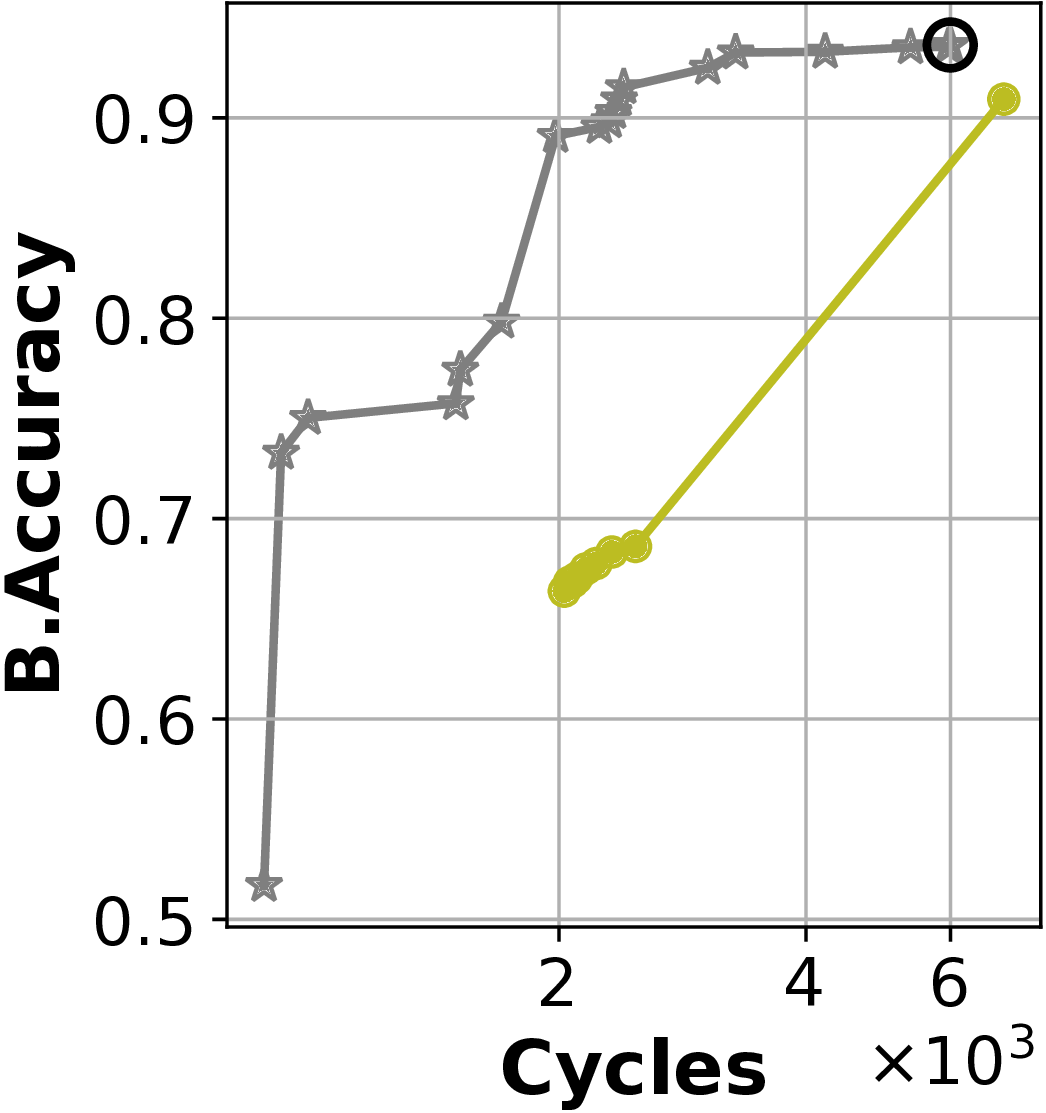}%
\label{fig:walk_adaptive}\subcaption{WALK}%
\end{subfigure}%
\caption{Variable-width networks results in terms of classification score versus average inference cycles.}
\label{fig:adaptive_vs_static}
\end{figure*}

Figure~\ref{fig:adaptive_vs_static} shows the results obtained converting one of the Pareto-optimal models from Figure~\ref{fig:cycles_optimization_results}, namely the one with the largest ``gain'', computed as in (\ref{eq:gain}), into a variable-width version. The graphs report the classification results versus the \textit{average} number of execution cycles over all the inputs in the test set. For each dataset, we report the static Pareto frontier (the same of Figure~\ref{fig:cycles_optimization_results}), highlighting the model that has been selected as a starting point for the realization of the adaptive network with a black circle. For better visualization, only a portion of the ranges of classification score and number of cycles is depicted. The yellow curves show the results obtained by the input-adaptive networks, where different points correspond to different settings of the confidence threshold ($T_h$), i.e., they represent different runtime operating modes. Specifically, higher scores correspond to larger $T_h$ values, which make the small model $M_s$ less ``confident'', causing more invocations of $M_l$.
For UniMiB-SHAR and WALK, $M_s$ is constructed using 50\% of the total channels, whereas for UCI HAPT and WISDM better results are obtained activating only 25\% of the channels for easy inputs.

The figure shows that, for three out of four datasets, the accuracy versus cycles trade-off obtained by the proposed adaptive system are competitive with those obtained with static models. In particular, on UCI HAPT, the adaptive solution obtains a significantly better trade-off, reducing the average number of cycles of up to 60\% for the same accuracy. This is due to the fact that the UCI HAPT test set contains a large number of samples from ``easy'' activity classes such as ``laying'' and ``walking'', which are confidently classified by $M_s$.

On UniMiB-SHAR and WISDM, the various operating modes obtained with the adaptive approach lie very close to the static Pareto frontier, except for those corresponding to the largest $T_h$ values, for which the overhead of two model invocations ($M_s$ first, and then $M_l$) becomes relevant.
It has to be noted that, for these two datasets, the class balance in the test set does not reflect the one typically found in the field. For example, in the UniMiB-SHAR test set, the total number of samples relative to different types of falls is superior to the number of ``walking'' samples, and activities such as ``sitting'' are not present at all. Consequently, while the adaptive network is competitive even in these unfavorable conditions, one can expect that if such a system is deployed in the field, its effectiveness could even increase, becoming more similar to the case of UCI HAPT.

\begin{figure}[ht]
    \centering
    \includegraphics[width=0.5\linewidth]{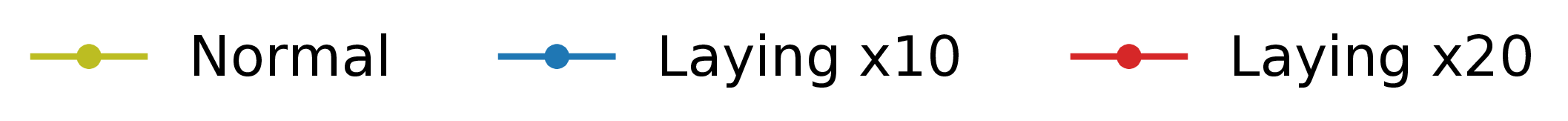}
    \includegraphics[width=0.5\linewidth]{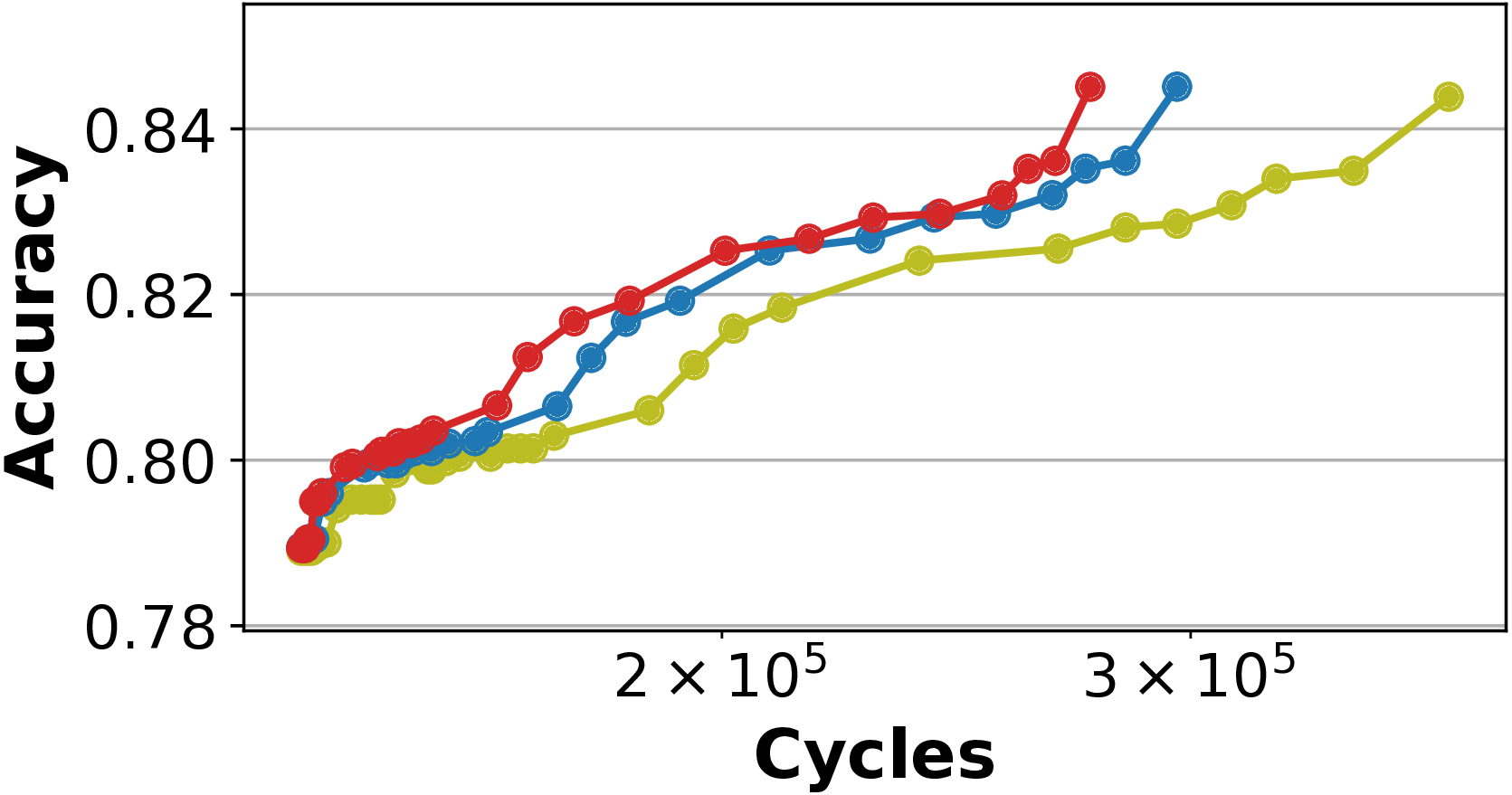}%
    \caption{Adaptive inference results for the UCI HAPT dataset with oversampling of the ``laying'' class.}
    \label{fig:hapt_slimmable_adaptive}
\end{figure}

One example of the impact of the data distribution is shown in Figure~\ref{fig:hapt_slimmable_adaptive}. We selected the UCI HAPT dataset for this experiment because it contains the ``laying'' class, which is particularly easy to classify. Namely, it is the class for which $M_s$ achieves the highest recall on the training set. The yellow curve is the same as in Figure~\ref{fig:adaptive_vs_static}, obtained on the ``standard'' UCI HAPT test set, in which approximately 15\% of the samples belong to the laying class. The blue and red curves, instead, are obtained executing the same adaptive network on modified test sets, in which laying samples have been replicated 10 and 20 times respectively, mimicking a scenario in which, for example, a wearable device is used also during sleep hours.
As shown, the trade-off between accuracy and average number of cycles per inference progressively improves with the number of laying samples, demonstrating the potential advantage of an adaptive model in cases in which most of the inputs are ``easy''.

The only dataset for which we could not obtain satisfying results with the proposed variable-width networks is WALK. This is because, for this dataset, most Pareto-optimal static networks are BNNs, including the one used as starting point to generate the adaptive model,
Conversely, for the other three datasets, we started from higher precision CNNs. Experimentally, we found that the class scores produced by BNNs are not representative of the actual probabilities of correct classification (i.e., they are not well calibrated~\cite{Guo2017}), making the score margin policy ineffective. Conversely, implementing the adaptive solution on top of a higher precision model would not be competitive either in this case, as the starting point would be too far from the static Pareto frontier. For this reason, we conclude that a variable-width approach is not competitive for this specific dataset. The investigation of adaptive inference solutions that can be applied effectively to BNNs will be a subject of our future work.

The main purpose of adaptive inference, however, is not to outperform a set of independent static models, but rather, as explained in Section~\ref{sec:adaptive}, to offer the flexibility of changing the accuracy versus inference cost trade-off at runtime with a low memory overhead. To demonstrate this advantage, Table~\ref{tab:adaptive} summarizes the total memory required to support the entire range of classification scores spanned by the variable-width networks of Figure~\ref{fig:adaptive_vs_static}. This memory requirement is compared with that of a naive system that covers the same range of scores using independent static models. Results are reported for the three public datasets, excluding WALK for the reasons discussed above.

\begin{table}[h]
\footnotesize
\begin{tabular}{cccccc}
\textbf{Dataset}   & \textbf{Score Range [\%]} & \textbf{Approach} &  \textbf{Cycles Range [$\cdot 10^5$]} & \textbf{N. of Points} & \textbf{Tot. Memory [kB]} \\ \hline
\multirow{2}{*}{UniMiB-SHAR} & \multirow{2}{*}{71:81 [BAcc.]} & Static & 1.6:3.1 & 5 & 50.6 \\
 & & Adaptive & 1.5:5.4 & 47 & 26.34 \\\hline
\multirow{2}{*}{UCI HAPT} & \multirow{2}{*}{78:83 [Acc.]} & Static & 3.06:6.71 & 4 & 39.4 \\
 & & Adaptive & 1.38:3.75 & 34 & 14.63 \\\hline
\multirow{2}{*}{WISDM} & \multirow{2}{*}{95:98 [F1]} & Static & 1.03:1.81 &  5 & 14.5\\
 & & Adaptive & 1.05:5.44 & 21 & 11.96 \\
\hline
\end{tabular}
\caption{Comparison between multiple static CNNs and a single input-adaptive one.}\label{tab:adaptive}
\end{table}

Precisely, the \textit{Score Range} column reports the minimum and maximum classification scores (Accuracy, B. Accuracy or F1, depending on the dataset) considered. The corresponding range of average clock cycles per classification is reported in the \textit{Cycles Range} column. The \textit{N. of Points} column reports the number of Pareto-optimal static models found by our grid search and mixed-precision optimization in the considered score range, and the number of Pareto-optimal operating modes of the adaptive system respectively. The latter have been found varying $T_h$ from 0 to 1 in 100 equally spaced steps on a log scale.
Lastly, the \textit{Tot. Memory} column reports the total memory occupation of the system: in the static case, this is just the sum of the independent models' sizes, whereas for the adaptive solution, it includes the overhead due to the switchable BatchNorm. 

For all datasets, resorting to an input-adaptive network allows to greatly increase the number of available operating points, in terms of classification score versus number of inference cycles (and consequently latency and energy consumption). Most importantly, the total memory needed to support these multiple working modes is reduced by 1.2x-2.7x.

\begin{figure*}[ht]
\centering
\includegraphics[width=.8\linewidth]{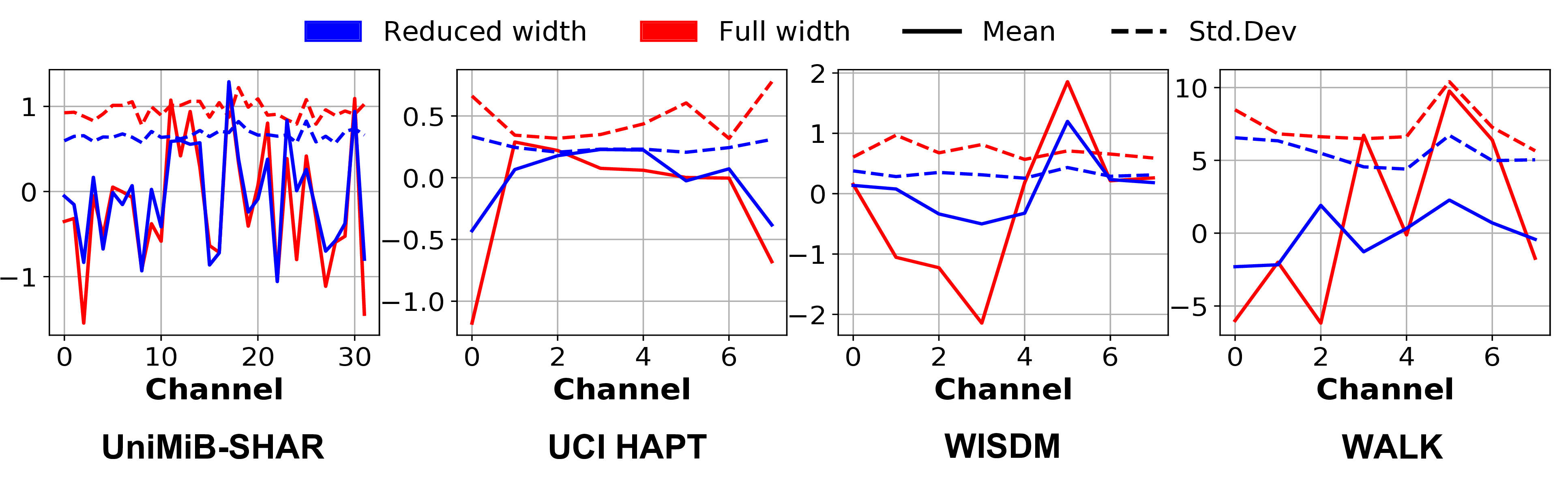}
\caption{Mean and Std.Dev.  of the last Batch Normalization layer at reduced- and full-width, for the proposed variable-width networks.}%
\label{fig:bn_comparison}%
\end{figure*}%

To confirm the benefits of switchable batch normalization, Figure~\ref{fig:bn_comparison} shows, as an example, the mean and standard deviation parameters of the last batchnorm layer for each variable-width model. Blue curves refer to the reduced-width model, and red ones to the full model. Only the channels that are active at reduced width are shown, for better visualization. As shown, the values at the two widths different significantly, in accordance to the findings of~\cite{yu2018slimmable}. Interestingly, the differences are more marked for the small models used for simple datasets such as WISDM and WALK. Switching the BN layer at runtime permits a compensation of these differences, at the cost of a limited memory overhead. 

\subsection{Comparison with state-of-the-art}

Table~\ref{tab:sota} compares our most accurate quantized CNN for each dataset with other DL approaches, and with a baseline implementation based on RFs, in terms of classification score and memory occupation. Concerning DL models, we selected the ones achieving the most accurate results that we could find in literature for each dataset, and we took model sizes and scores directly from the original papers. 
In contrast, 
none of the state-of-the-art papers based on classic ML algorithms (e.g., k-NN, RF, etc.) that we analyzed reports the memory occupation of the models, meaning that we 
can only compare with them in terms of classification score, discussing complexity aspects qualititatively, as done in the following. 
Therefore, we decided to use the efficient RF implementation of~\cite{daghero2021ultracompact} in order to also have a complete and fair complexity comparison with a representative family of non-deep classifiers. 

We selected RFs as baselines due to their limited memory requirements and good accuracy, which makes them popular for embedded deployment~\cite{stsensor}. For each dataset, we performed a grid search on the depth and number of trees in the RF, using the same train/test split of our CNNs. Additionally, we tested both RFs that take as input the raw acceleration and gyroscope data, and versions that process the features proposed in~\cite{comparison_hapt_rf, comparison_wisdm_rf,micucci2017unimib}, most of which have been implemented on the target MCU using the library of~\cite{pulp_dwt}. We then selected the most accurate combination of hyper-parameters and input type for each dataset. 

\begin{table}[h]
\begin{adjustbox}{max width=\linewidth}
\begin{tabular}{llllllll}
                 & \multicolumn{2}{l}{\textbf{Ours (Best)}}       & \multicolumn{3}{l}{\textbf{Previous DL Works}}  & \multicolumn{2}{l}{\textbf{Baseline RF}}            \\\hline\hline
\textbf{Dataset} & \textbf{Score [\%]}  & \textbf{Mem. [kB]}  & \textbf{Paper} & \textbf{Score [\%]} & \textbf{Mem. [kB]} & \textbf{Score [\%]} & \textbf{Mem. [kB]} \\\hline
UniMiB & 86.24/90.66 [BAcc./F1] & 23.16 & \cite{comparison_wisdm_unimib_dnn} & n.a./77.8 [BAcc./F1] & 5800 & 58.05/65.61 [BAcc./F1] & 202.17 \\ %
UCI HAPT    & 85.63 [Acc.] & 7.53  & \cite{comparison_hapt_dnn}*        & 97.98 [Acc.] & 17939 & 74.16 [Acc.] & 51.71  \\
WISDM       & 98.9/98.81 [F1/Acc.] & 6.21  & \cite{comparison_wisdm_unimib_dnn} & 98.8/n.a. [F1/Acc.] & 1640  & 93.91/94.16 [F1/Acc.] & 255.74 \\
WALK        & 95.74 [BAcc.] & 1.64  & \cite{daghero2021ultracompact}     & 94.63 [BAcc.] &  0.9  & 91.86 [BAcc.] & 8.26   \\\hline
\end{tabular}
\end{adjustbox}
\caption{Comparison with state-of-the-art Deep Learning (DL) models and with a baseline implementation with Random Forests (RF). The classification score is the same used in previous graphs and tables except for UniMiB and WISDM, where we report two metrics to fairly compare with~\cite{comparison_wisdm_unimib_dnn} and~\cite{mobiact}. (*: 8-class classification).}\label{tab:sota} 

\end{table}

On three of the four datasets (UniMiB-SHAR, WISDM and WALK), our CNNs outperform previous DL solutions in terms of classification scores. 
Note that for UniMiB we report both the B. Accuracy (the original metric proposed for this dataset~\cite{micucci2017unimib}) and the F1-Score, which is the only metric reported by the DL work of~\cite{comparison_wisdm_unimib_dnn}. On UCI HAPT, the authors of~\cite{comparison_hapt_dnn} obtain a higher score than ours, but a direct comparison is impossible in this case, since they focus on an easier version of the problem (8-class versus 12-class). Nonetheless, we are still able to achieve an acceptable accuracy on a harder task, with far less parameters.

Most importantly, on three datasets, we achieve striking memory reductions with respect to previous deep models. Namely, we reduce the model size by 2400x on UCI HAPT, and $\approx$250x (with higher scores) on UniMiB-SHAR and WISDM. The only exception is WALK, for which the previous best result was obtained in~\cite{daghero2021ultracompact} with a very compact BNN, smaller but 1.1\% less accurate than the best 4-bit CNN found in this work. This memory reduction is fundamental for embedded deployment. For instance, if we assume a maximum memory constraint of 256kB, i.e., 50\% of the total main memory available in the target MCU~\cite{quentin}, all our most accurate CNNs can be easily deployed. On the contrary, none of the previous DL approaches fits, except for the BNN of~\cite{daghero2021ultracompact}. Note that Table~\ref{tab:sota} only reports the most accurate models from previous works. However, even less accurate models proposed in those papers are still significantly bigger than ours. For example, the smallest model proposed in~\cite{comparison_wisdm_unimib_dnn} achieves a F1-score of 72.80\% on UniMiB and 96.30\% on WISDM, while still requiring 0.88M and 0.76M of memory respectively.
These considerations are valid if the deployment target is a tightly constrained device of the class discussed in Section~\ref{sec:nodes}, i.e., a system based on a low-power MCU, such as those found in cheap fitness-trackers. There also exist much more powerful devices for which HAR models are of interest, such as full-fledged smartwatches, which are based on Cortex-A-like processors and equipped with 100s of MB of memory. Clearly, those allow the deployment of most state-of-the-art models discussed above. However, our CNNs remain of high interest even for those devices, as their smaller size translates into a much faster and energy-efficient inference (as further detailed in Section~\ref{sec:deployment}), while retaining a comparable accuracy.

Our CNNs also outperform the best RF-based implementation fitting on the target MCU for all datasets, while requiring significantly less memory for deployment (from 5x to 41x depending on the benchmark). On UniMiB-SHAR and WALK, we outperform all classic models from previous literature too (see Table~\ref{table:related}). In particular, for the UniMiB-SHAR AF-17 task, \cite{micucci2017unimib} reports a k-NN classifier and a RF achieving respectively 82.86\% and 81.48\% balanced accuracy, both 
lower than our best CNN. Moreover, those models are not compatible with MCU deployment, since k-NN requires on-device storage of the features of \textit{all training data} (at least 7MB of memory based on the number of features reported in the paper and assuming a float representation). Similarly, the best RF from~\cite{micucci2017unimib} includes 300 trees. The total memory occupation of the model is not reported, and making an estimate is impossible since the number of nodes and/or depth of the trees is also not mentioned in the paper. However, qualitatively, it is reasonable to assume that the memory required would not be compatible with our constraints, since the most accurate RF that we found occupies 202.17kB while including just 5 trees of depth 16.
In~\cite{comparison_hapt_rf}, the authors achieve 88\% accuracy on the 12-class UCI HAPT (slightly higher than ours) with a RF-based classifier, but without specifying the hyper-parameters.
Lastly, a 99.8\% accuracy score is obtained in~\cite{mobiact} on WISDM, again using k-NN. This result outperforms our models by less than 1\% (our best CNN's accuracy is 98.81\%). However, note that standard accuracy is not as reliable as the F1 Score for a highly imbalanced dataset such as WISDM. Moreover, the k-NN solution of~\cite{mobiact} is again hardly compatible with MCU deployment, as it requires at least 12.6 MB for storing all training data on-device.
These and other scores comparisons with DL and classic models are reported in Table~\ref{table:related}.

\subsection{Detailed Deployment Results}\label{sec:deployment}

Table~\ref{tab:deployment} reports the detailed deployment metrics of some of the Pareto-optimal CNNs identified in our work. Specifically, we report the memory occupation, inference latency and energy consumption of three models per dataset, together with the corresponding classification scores. We select the two extremes of our exploration, i.e., the most accurate and the smallest model. Moreover, we also report the results of an intermediate CNN, namely the smallest one that incurs a score drop $<5\%$ with respect to the most accurate (Max - 5\% in the table). All results refer to Quentin running at 205.1 MHz, with a supply voltage of 0.54 V, and assuming that the MCU is power gated after the end of an inference.

\begin{table}[h]
\footnotesize
\begin{tabular}{cllclll}
\textbf{Dataset}   & \textbf{Config}    & \textbf{Score [\%]} & \textbf{Metric}      & \textbf{Memory [kB]} & \textbf{Energy [$\mu J$]} & \textbf{Latency [$ms$]} \\ \hline \hline
\multirow{3}{*}{UniMiB-SHAR}    & Min       &  26.24 & \multirow{3}{*}{B.Acc.}       &  0.41           &   0.34              & 0.09             \\ 
                                & Max - 5\% & 81.56 &      & 9.28            & 29.33                 & 7.7            \\ 
                                & Max       & 86.24 &      &  23.17          & 61.59                    & 16.2             \\ \hline 
\multirow{3}{*}{UCI HAPT}       & Min       & 44.53   & \multirow{3}{*}{Acc.}      & 0.42            & 0.44                  & 0.12               \\ 
                                & Max - 5\% & 83.18  &     & 4.6            & 30.52                  & 8.01              \\ 
                                & Max       & 85.63   &    & 7.54           & 23.27                 & 6.11             \\ \hline  
\multirow{3}{*}{WISDM}          & Min       & 67.6   & \multirow{3}{*}{F1}          & 0.22            &  0.3                 & 0.08                \\ 
                                & Max - 5\% &  94.74  &    & 1.27            & 4.09                  & 1.07           \\ 
                                & Max       & 98.9 &      & 6.22           & 15.94                 & 4.19          \\ \hline 
\multirow{3}{*}{WALK}           & Min       & 78.13 & \multirow{3}{*}{B.Acc.}             & 0.05            & 0.03                 & 0.009               \\ 
                                & Max - 5\% &  91.81  &     & 0.18 & 0.05 &  0.016\\ 
                                & Max       & 95.74  &     & 1.65            & 0.67                  & 0.18             \\ \hline 
\end{tabular}\caption{Detailed deployment results at different trade off points in terms of score versus memory.}\label{tab:deployment}
\end{table}\label{table:results}

The table shows that, thanks to our extensive design space exploration, we are able to span multiple orders of magnitude in memory, energy consumption and latency. Moreover, even our most accurate models are suitable for real-time inference, since the total latency is at most 16ms (for UniMiB-SHAR), i.e., much smaller than the length of the input signals time-windows used by the four considered datasets (see Table~\ref{table:related}).

The energy consumption of our models is also extremely low. For example, our most consuming CNN for WISDM would deplete a
small fitness tracker Li-Polymer battery with a capacity of 30mAh@3.7V~\cite{battery}
after more than \textit{7 years} of continuous on-device activity recognition, with one inference every 10s, as required by the input windowing of~\cite{wisdm}. Clearly, this does not correspond to the real battery life-time, since it ignores all other sources of power consumption (other tasks running on the MCU, and external peripherals such as sensors, displays, etc.), as well as all battery non-idealities and power conversion losses. However, it provides a quantitative idea of the efficiency of our models. 

Lastly, an interesting observation is that our ``Max - 5\%'' CNNs often achieve a very significant latency, energy and memory reduction compared to ``Max'' models, at the cost of a small accuracy drop. For example, for the WALK dataset, the ``Max - 5\%'' CNN is 10x more efficient and smaller than ``Max'', while still achieving a very good accuracy of 91.8\%.

\begin{figure*}[ht]
\centering
\includegraphics[width=\linewidth]{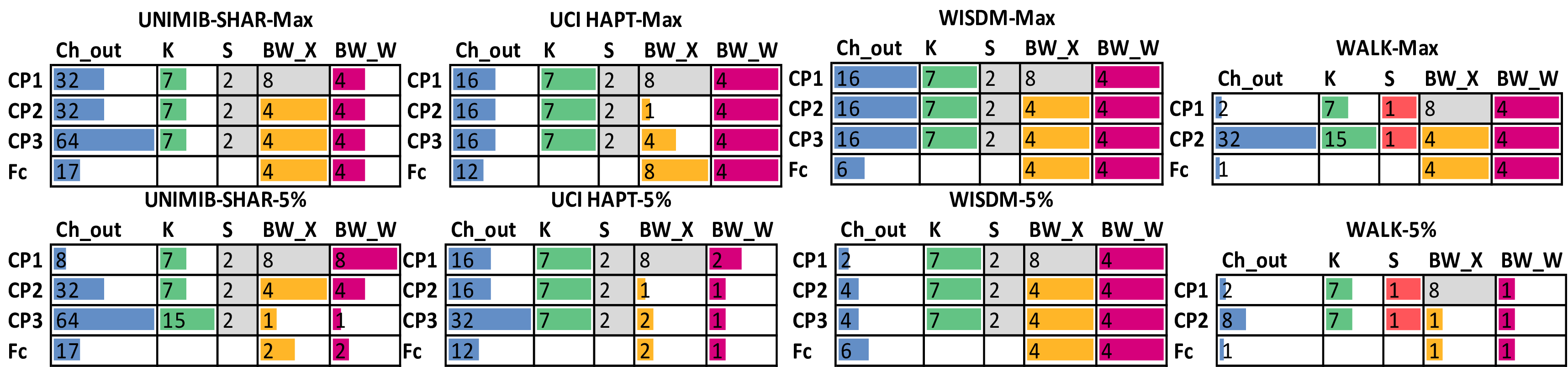}
\caption{Architectures of the ``Max'' (1st row) and ``Max-5\%'' (2nd row) CNNs depolyed on the taget MCU. Legend: CP $=$ Convolution + Pooling block, FC $=$ fully-connected layer. C\_out $=$ n. of output channels, K $=$ kernel size, S $=$ pooling stride, BW\_X $=$ input bit-width, BW\_W $=$ weights bit-width.}%
\label{fig:best_comparison}%
\end{figure*}%

A detailed view of the ``Max'' and ``Max-5\%'' model architectures is shown in Figure~\ref{fig:best_comparison}.
Specifically, we show the main hyperparameters of each convolution+pooling block (CP) and fully connected layer (Fc), reporting the output channels, the kernel size, the pooling stride and the quantization bit-width used for the input activations and weights. Note that the pooling stride is not explored for all datasets, and therefore it is greyed out, similarly to the input bit-width of the first CP block which is always fixed at 8-bit. Also, we do not show the ``Min'' models since these are always equal to the smallest-size extreme of the search space, i.e., for all layers: $C_{out} =2$, $K=7$, $BW\_X = BW\_W = 1$ and for the Walk model, $S=4$.

The figure provides some interesting insights on the found Pareto-optimal models. For instance, it often holds that $BW\_X \ge BW\_W$, confirming the well-known observation that good models often need a higher precision for activations~\cite{hubara,edmips}. Similarly, there is a general trend towards increasing $C_{out}$ in deeper layers, which matches a popular rule-of-thumb of manually designed networks. %
Comparing the architectures for different datasets, we also notice that different mechanisms are effective to reduce model size while preserving accuracy (i.e., to go from ``Max'' to ``Max-5\%'' models). Namely, for UniMiB-SHAR and UCI HAPT, the main difference between the two models is a reduction in the quantization bit-width, combined with an \textit{increase} of the geometrical shape of the last convolutional layer, whereas for WISDM quantization is identical, and the last layer geometry is altered by \textit{reducing} $C_{out}$. Together, these results demonstrate the importance of exploring both network architecture and quantization precision in order to obtain good CNNs for HAR.
\section{Conclusion}\label{sec:conclusion}

In this paper, we have described an extensive exploration of efficient 1D CNN architectures for HAR on MCUs. Namely, we have combined architectural optimization with sub-byte and mixed-precision quantization, in order to obtain good trade-offs between classification score and memory occupation, which is typically the main factor liming the deployability of deep learning models on embedded devices. Orthogonally, we have also explored adaptive inference using variable-width CNNs as a way to provide higher runtime flexibility to the system, allowing an easy switch between multiple operating modes with a limited memory overhead.

Targeting four different HAR datasets, we have demonstrated that sub-byte precision networks are a competitive alternative to shallow learning approaches and higher precision CNNs. On three of the benchmarks, our networks outperform all previous deep learning approaches, while requiring orders of magnitude less memory. The few classifiers (shallow or deep) that achieve better results are not deployable on low-power MCUs. Moreover, we have shown that adaptive inference offers an effective solution to obtain a flexible and runtime-reconfigurable system, and that even our most complex CNNs are efficient enough to allow years of continuous HAR on a small battery.

\bibliographystyle{ACM-Reference-Format}
\bibliography{library}

\end{document}